\documentclass{article}

\usepackage{microtype}
\usepackage{graphicx}
\usepackage{subfigure}
\usepackage{booktabs} 

\usepackage{hyperref}

\usepackage[normalem]{ulem}

\usepackage[accepted]{icml2024}

\usepackage{amsmath}
\usepackage{amssymb}
\usepackage{mathtools}
\usepackage{amsthm}
\usepackage{enumitem}

\usepackage{multirow,multicol}
\usepackage{CJKutf8}

\usepackage[capitalize,noabbrev]{cleveref}

\theoremstyle{plain}

\theoremstyle{definition}

\theoremstyle{remark}

\usepackage[textsize=tiny]{todonotes}

\icmltitlerunning{Linear Correlation in LM's Compositional Generalization and Hallucination}

\begin{document}

\twocolumn[
\icmltitle{Linear Correlation in LM's Compositional Generalization and Hallucination}



\icmlsetsymbol{equal}{*}

\begin{icmlauthorlist}
\icmlauthor{Letian Peng}{ucsd}
\icmlauthor{Chenyang An}{ucsd}
\icmlauthor{Shibo Hao}{ucsd}
\icmlauthor{Chengyu Dong}{ucsd}
\icmlauthor{Jingbo Shang}{ucsd}
\end{icmlauthorlist}

\icmlaffiliation{ucsd}{University of California, San Diego}

\icmlcorrespondingauthor{Jingbo Shang}{jshang@ucsd.edu}

\icmlkeywords{Machine Learning, ICML}

\vskip 0.3in
]



\printAffiliationsAndNotice{}  

\begin{abstract}
The generalization of language models (LMs) is undergoing active debates, contrasting their potential for general intelligence with their struggles with basic knowledge composition (e.g., reverse/transition curse).
This paper uncovers the phenomenon of \emph{linear correlations} in LMs during knowledge composition. 
For explanation, there exists a linear transformation between certain related knowledge that maps the next token prediction logits from one prompt to another, e.g., ``\textit{X lives in the city of}''$\rightarrow$ ``\textit{X lives in the country of}'' for every given \textit{X}. 
This mirrors the linearity in human knowledge composition, such as \textit{Paris}$\rightarrow$ \textit{France}. 
Our findings indicate that the linear transformation is 1) resilient to large-scale fine-tuning, 2) generalizing updated knowledge when aligned with real-world relationships, 3) but causing hallucinations when it deviates. 
Empirical results suggest that linear correlation can serve as a potential identifier of LM's generalization. 
Finally, we show such linear correlations can be learned with a single feedforward network and pre-trained vocabulary representations, indicating LM generalization heavily relies on the latter. \footnote{Code: \href{https://github.com/KomeijiForce/LinCorr}{https://github.com/KomeijiForce/LinCorr}}

\end{abstract}

\section{Introduction}

What knowledge do language models (LMs) learn beyond memorizing the training data? The generalization ability of LMs is undergoing an active debate. Optimists claim that LMs might have the capability in entirely novel tasks with their emergent behavior~\citep{emergent_ability} by scaling-up parameters, while pessimists argue that LMs struggle with composing simple knowledge~\citep{compositional_limitation,alg_compositional_limitation}, such as reverse or transition curses claiming that LMs cannot even simply compose knowledge by reversing or transiting~\citep{reverse_curse,curse_theory}.

\begin{figure*}
    \centering
    \includegraphics[width=0.99\linewidth]{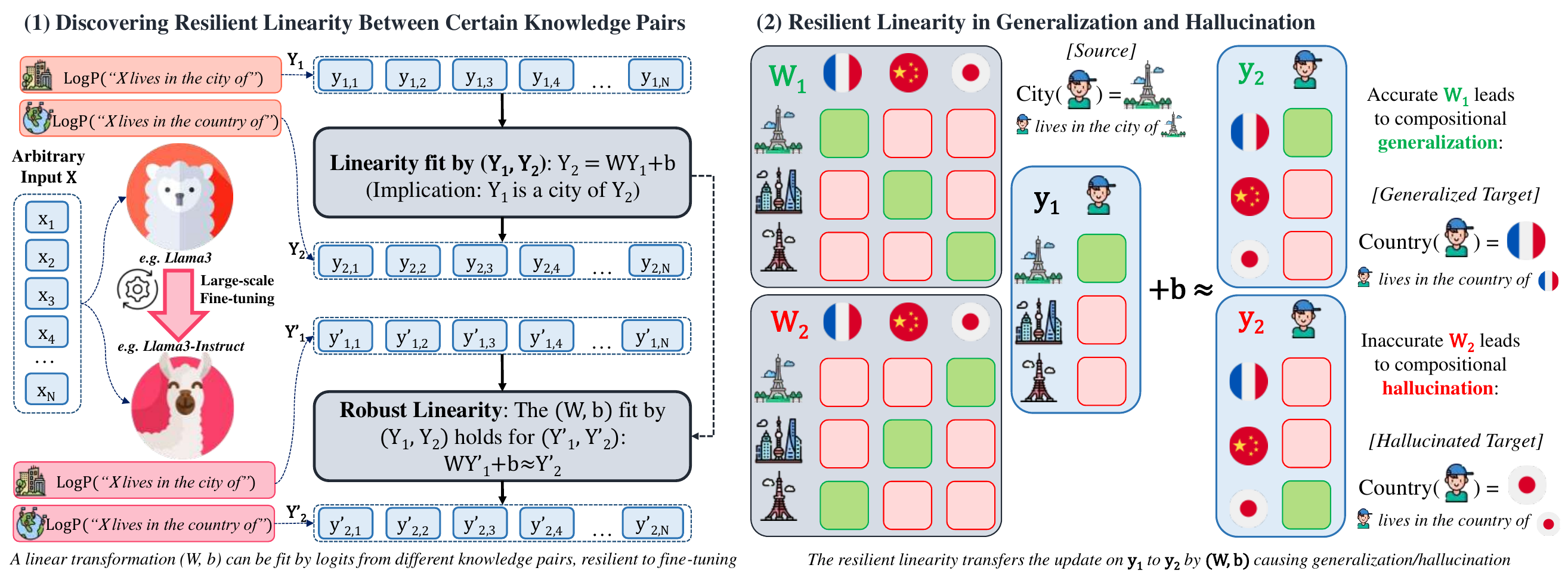}
    \vspace{-3mm}
    \caption{
    Demonstration of our main discoveries. 1) We can fit a linear transformation between the output of source and target knowledge prompts, which is resilient against fine-tuning. 2) Updating the source knowledge will generalize to the target one via resilient linearity, causing compositional generalization/hallucination.
    }
    \vspace{-3mm}
    \label{fig:intro}
\end{figure*}

{While macroscopically investigating how skills emerge in language models remains challenging}, we can gain microscopical insight from the generalization behavior on the smallest learning unit, next token prediction (NTP). We unveil an interesting linear correlation between logits of related NTPs, such as \textit{City}$\rightarrow$\textit{Country}, from the \textbf{source knowledge} like logits of $F_{\textit{City}}(X)=$ NTP$(``\textit{X lives in the city of''})$ to the \textbf{target knowledge} like logits of $F_{\textit{Country}}(X)=$ NTP$($\textit{``X lives in the country of''}$)$. 
Between logits in knowledge subdomains (e.g., $\{\textit{Paris}, \textit{Shanghai}, \cdots\}$ for $F_{\textit{City}}(X)$), we can fit a linear transformation $(W, b)$ that well approximates $F_{\textit{Country}}(X) = W \cdot F_{\textit{City}}(X) + b$ for any $X$ as the input. 
To fit the transformation, we sample numerous output logits from prompts with arbitrary inputs $X$s as shown in Figure~\ref{fig:intro}. Then, $(W,b)$ is fitted with partial logit pairs and tested on the rest. {The Pearson correlation coefficients for evaluation reflects the inherent relations of knowledge in the real world, with high correlations in cases like \textit{City}$\rightarrow$\textit{Country} and low correlations in cases like \textit{City}$\rightarrow$\textit{Gender}}.

{Examining $W$}, we find that its weights mirror the linearity in the knowledge composition of humans. In the \textit{City}$\rightarrow$\textit{Country} case, the $W$ assigns high weights to real-world (City, Country) pairs such as \textit{Paris}$\rightarrow$\textit{France}. In other words, probability $P(F_{\textit{Country}}(X)=\textrm{\textit{France}})$ is correlated with $P({F_\textit{City}}(X)=\textrm{\textit{Paris}})$. However, there also exists counterfactual weights learned in $W$, for instance, the weight fit in $W$ for (\textit{Indianapolis}, \textit{India}) is much higher than the correct (\textit{Indianapolis}, \textit{USA}). {We say $W$ is \textit{precise} when $W$ assigns high weights for the correct knowledge pairs}.
$W$'s precision is generally low for knowledge pairs with low correlations, but a high linear correlation also does not guarantee high precision.
{This motivates us to explore the connection between 1) such linear correlations, 2) $W$'s precision, and 3) LM's compositional generalization}.
{Importantly, if the same $W$ and $b$ also fit the parameter updates after gradient propagation, then learning source knowledge will simultaneously update the target knowledge}.


We begin with one-step parameter updates, fine-tune the LM with a piece of source knowledge, and then check the gradients on the source and target knowledge. When the linear correlation between the source and target knowledge is high, we find $W$ capable of estimating the gradients on the target knowledge based on the source gradient. We then extend the comparison to LMs before and after large-scale post-training, which shows $W$ fitted before post-training to {retain} the estimation ability for the LM after post-training. Thus, $W$ between highly correlated knowledge is found resilient against gradient propagation, which consistently plays an important role in generalization. 



To validate the important role of linear correlation in LM generalization, we test the generalization effect between source and target knowledge with different levels of correlation intensity and $W$ precision. Our study shows that a successful generalization {for a simultaneous knowledge update between source and target} requires high correlation intensity and $W$ precision. This implies that LMs struggle to generalize their predictions in a non-linear manner, explaining why simple fine-tuning cannot efficiently edit LMs~\citep{cohen2024evaluating}. {When the Pearson coefficient is high and $W$ is imprecise}, the resilient linear correlation will consequently lead to compositional hallucination. For instance, learning $P(City(X)=\textit{Indianapolis})$ unfortunately generalizes to $P(Country(X)=\textit{India})$. {Our linear correlation reflects the occurrence of such hallucinations before fine-tuning, demonstrating its utility in diagnosing potential faults in the knowledge composition of LMs.}

Finally, we explore the linear correlation’s origin and hypothesize that vocabulary representations are key. Even when we remove the LM’s complex internals (position embeddings, self-attention, etc.) and use only a mean-pooling layer plus a single feedforward network, the model still learns to compose knowledge from few paired texts (e.g., $F_{\textit{City}}=\textit{Paris}$ paired with $F_{\textit{Country}}=\textit{France}$). The simplified archecture shows similar generalization performance as the original Transformer. However, altering lexical mappings (e.g., \textit{Paris}$\rightarrow$\textit{Japan}) disrupts this ability, underscoring the critical role of vocabulary representations.

Our contributions are presented as follows,

\begin{itemize}[nosep,leftmargin=*]
    \item We unveil the linear correlation between the LM's output logits for related knowledge.
    \item We find such linear correlation existing between gradients and resilient against training, which connects it to compositional generalization and hallucination of LMs. 
    \item We attribute the formation of the linear correlation between NTPs to the vocabulary representations.
\end{itemize}

\section{Related Works}

\subsection{Language Model Interpretation}
Language models (LMs) ~\citep{achiam2023gpt4,team2024gemma,olmo,dubey2024llama3} are gaining widespread attention across various fields due to their strong performance on a variety of tasks, like reasoning and knowledge retrieval. However, the black-box nature of (neural) LMs hinders human's understanding of their working mechanism. Various methods have been developed to interpret LM behavior by analyzing its parameters and intermediate representations. Several works suggest that LMs store knowledge inside the feedforward layers~\citep{key_value,knowledge_neuron,locate_edit_fact}, which are used in a key-value matching manner to map inputs into related knowledge~\citep{ffn_vocab}. Some parameters are also found to perform certain relational transformations for the LM~\cite{fv,linguistic_region}, known as the task representations~\citep{task_representation}.
For certain subsets of relations, LMs have been unexpectedly found to encode knowledge in a linear manner~\citep{linear_relation}, suggesting a potential role of linearity in their understanding of relational structures. However, it remains unknown how the LM understands the transformation between relations. Our work shows the linearity between the output from several relation pairs given the same input.

\subsection{Model Generalization}

The power of modern deep neural networks lies in their remarkable ability to generalize effectively to unseen inputs. However, the exact mechanisms through which these models achieve generalization remain poorly understood. For instance, in the context of knowledge editing, numerous research studies have observed that standard fine-tuning methods for updating knowledge often struggle to meet critical objectives simultaneously \cite{onoe2023can, hoelscher2023detecting, meng2022locating, gupta2023editing}. On one hand, they fail to prevent unintended modifications to unrelated knowledge. On the other hand, they frequently fall short of ensuring that logical deductions based on the updated knowledge are properly incorporated \cite{cohen2024evaluating, zhong2023mquake}. 
Previous research has proposed various metrics and methods to measure and predict generalization in deep neural networks. However, these approaches don't cover the perspective of correlation in model generalization proposed in our work \cite{yu2022predictingoutofdistributionerrorprojection, garg2022leveragingunlabeleddatapredict, kang2024learningdynamicsrevealgeneralization}.

\subsection{Hallucination Detection}

Hallucination remains one of the most significant challenges in the deployment of language models (LMs) \cite{zhang2023sirenssongaiocean, Huang_2024}. Numerous studies have explored approaches to predict and mitigate this issue. For instance, some prior works utilize trained classifiers to identify hallucinations \cite{jiang2024largelanguagemodelshallucination, quevedo2024detectinghallucinationslargelanguage, chen2024hallucinationdetectionrobustlydiscerning}. Another method involves detecting hallucinations by clustering semantically similar responses and calculating entropy across these clusters \cite{farquhar2024detecting}. Additionally, the MIND framework has been proposed to exploit the internal states of LMs during inference, enabling real-time hallucination detection \cite{su2024unsupervisedrealtimehallucinationdetection}. Moreover, formal methods guided by iterative prompting have been employed to dehallucinate LM outputs \cite{jha2023dehallucinating}. RAG has also been used to detect and correct hallucinations in LM \cite{mishra2024finegrainedhallucinationdetectionediting}. Our study presents an innovative approach to predicting hallucinations, different from existing methodologies, by leveraging the correlation.

\section{Discovering Linear Correlation }


\subsection{Preliminary and Motivation}
\paragraph{Next Token Prediction.} Neural language models have been scaled up to numerous parameters but can still be understood as a mapping function among vocabulary representations $V\in \mathbb{R}^{\textrm{\#Vocab}\times d}$.
We denote the embedding of the word \textit{X} as $V_\textit{X} \in \mathbb{R}^{d}$.
For an input word sequence, such as ``\emph{X lives in the city of}'', the embeddings of the involved words will be processed with other components in the LM $\theta_{\neg V}$ (positional embedding, self-attention networks, etc.) to encode the input context as $C=F([V_\textit{X}, \cdots, V_{\textit{of}}])\in \mathbb{R}^d$. 
Most\footnote{In Appendix~\ref{apdx:whole}, we empirically show our conclusion also holds for an parameter untied LM - Mistral~\citep{jiang2023mistral}}, if not all, LMs tie the input and output vocabulary embeddings together~\citep{vocab_tie} to use the dot product $C \cdot V_{\textit{Y}}$ as the logit of \textit{Y} for the next token prediction. Finally, the vocabulary-wise dot products are normalized by a softmax layer to represent the probability of a certain token (\textit{Y} for example).\footnote{We omit the discussion of potential bias terms, multiple token input for simplification and reading fluency.}

\begin{equation}
\small
    P_{\theta_{\neg V}}(\textit{Y}|[V_{\textit{X}}, V_{\textit{lives}}, \cdots, V_{\textit{of}}]) = \frac{e^{C\cdot V_{\textit{Y}}}}{\sum_{Z\in \textrm{Vocab}} e^{C\cdot V_{\textit{Z}}}}
\end{equation}
For a subset of all possible sequences that follow the template ``\emph{X lives in the city of}'' and takes arbitrary \textit{X} as the input, we can view template representations $[V_{\textit{lives}}, \cdots, V_{\textit{of}}]$ as constant to map a variable $X$ ($V_X$) with the \textit{City} relation. 

\begin{equation}
\small
    P_{\theta_{\neg V}}(\textit{Y}|[V_{\textit{X}}, V_{\textit{lives}}, \cdots, V_{\textit{of}}]) = P_{\theta_{\neg V}, [V_{\textrm{lives}}, \cdots, V_{\textrm{of}}]}(\textit{Y}|V_{\textit{X}})
\end{equation}

Here, the encoding function $F_{\textrm{City}} = F(\cdot|[V_{\textit{lives}}, \cdots, V_{\textit{of}}])$ (subscript \textit{City} denotes the semantics of constant representations) affects the final probabilistic distribution by mapping $V_{X}$ to $C$ near vocabulary embeddings of cities, such as $V_{\textit{Paris}}$, $V_{\textit{Shanghai}}$, $V_{\textit{Tokyo}}$. 

\paragraph{Motivation: Linearity in Relation.} Some knowledge like $F_{\textrm{CityToCountry}}$ (``\textit{X is a city in the country of}'') are found linear~\citep{linear_relation} between vocabulary representations, which means $F$ can be well approximated by $(W, b)$ s.t. $C = WV + b$. While not all mappings have such an interesting property, this phenomenon indicates the potential for LMs to compose knowledge in their parameters.

\paragraph{Knowledge Composition.} There exists compositional relations between knowledge such as $F_{\textrm{Country}}$ (``\textit{X lives in the country of}'') can be composed by other relations as $F_{\textrm{CityToCountry}}(F_{\textrm{City}})$ since one's residential city \textbf{(source knowledge)} indicates one's residential country \textbf{(target knowledge)}. Suppose the LM applies $F_{\textrm{City}}(V_{X})$ to map $V_{X}$ close to a city embedding like $V_{\textit{Paris}}$), then may the LM learn $(W, b)$ inside parameters and perform $F_{\textrm{Country}}(V_{X}) = F_{\textrm{CityToCountry}}(F_{\textrm{City}}(V_{X}))=WV_{\textit{Paris}}+b=V_{\textit{France}}$? While the hypothesis can be made for non-linear relations in the composition as well, we emphasize the linearity as it corresponds to the key-value matching~\cite{key_value} behavior of Transformers. The linear transformation can be simply performed by a feedforward network activated by self-attention. 

\begin{figure}
    \centering
    \includegraphics[width=\linewidth]{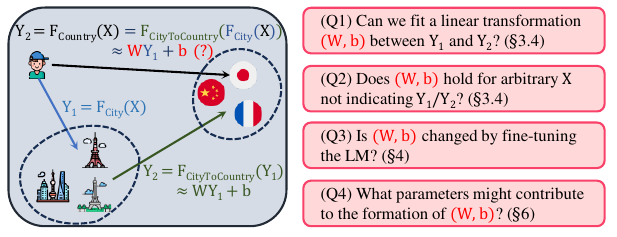}
    \vspace{-5mm}
    \caption{
    Our hypothesis and questions about how LMs compose knowledge by learning $(W, b)$.}
    \vspace{-2mm}
    \label{fig:questions}
\end{figure}

Motivated by the potential role of linearity in compositional knowledge, we conduct experiments to validate the hypothesis that LMs learn such linear transformation inside the parameters to compose knowledge. The roadmap of our exploration is presented in Figure~\ref{fig:questions}, with questions we will answer in the following sections. We will demonstrate that

\begin{itemize}[nosep,leftmargin=*]
    \item Such $(W, b)$ exists for logits prompted from certain related knowledge pairs, which is applicable to arbitrary inputs, not necessarily indicating a known output ($\S$~\ref{subsec:harvest}).
    \item Such linearity stays resilient against large-scale fine-tuning, which guarantees the LM's generalization to compositional knowledge ($\S$~\ref{sec:robust}).
    \item Such linearity can be highly attributed to the vocabulary representations. ($\S$~\ref{sec:building}).
\end{itemize}

\subsection{Method and Evaluation}

We search for the potential linear transformation between pairs of source and target knowledge. Continuing with the $(F_{\textrm{City}}, F_{\textrm{Country}})$ example, the transformation will be established between $C_{\textrm{City}, \textit{X}} = F_{\textrm{City}}(V_{\textit{X}})$ and $C_{\textrm{Country}, \textit{X}} =F_{\textrm{Country}}(V_{\textit{X}})$. We then decode the two representations by the LM head to produce logits $\textrm{LogP}_{\textrm{City}, \textit{X}}$ and $\textrm{LogP}_{\textrm{Country}, \textit{X}}$ both in shape $\mathbb{R}^{\textrm{\#Vocab}}$. 

\begin{equation}
\small
\begin{aligned}
    \textrm{LogP}_{\textrm{City}, \textit{X}}=C_{\textrm{City}, \textit{X}}\cdot V; \textrm{LogP}_{\textrm{Country}, \textit{X}}=C_{\textrm{Country}, \textit{X}}\cdot V
\end{aligned}
\end{equation}

As the dot product with $V$ is linear, the potential linearity holds after the transformation. We can calculate $W\in \mathbb{R}^{\textrm{\#Vocab}\times \textrm{\#Vocab}}$ and $b\in \mathbb{R}^{\textrm{\#Vocab}}$ for the transformation between logits. We learn $(W, b)$ for logit transformation (rather than hidden state) to improve the interpretability of the fitted $W$. For example, a high weight in $W_{(\textrm{France}, \textrm{Paris})}$ indicates a correct understanding of knowledge composition.

In practice, only a subdomain\footnote{General subdomain size is $\sim100$, as listed in Appendix~\ref{apdx:prompts}.} $D$ of the LM's large vocabulary is meaningful for the predicted logits, such as $D_{\textrm{City}}=$ \{\textit{Paris}, \textit{Shanghai}, \textit{Tokyo}, $\cdots$\} for $\textrm{LogP}_{\textrm{City}}$ and $D_{\textrm{Country}}=$ \{\textit{France}, \textit{China}, \textit{Japan}, $\cdots$\} for $\textrm{LogP}_{\textrm{Country}}$. Thus, we are more interested in the submatrix of $W$ for these meaningful words. Our main experiments will focus on those values in $W$ representing the linear transformation $W_{(D_{\textrm{City}}, D_{\textrm{Country}})}$ between such output subdomains. The specific procedure to build such subdomains is presented in Appendix~\ref{apdx:subdomain}.

Based on the prior discussion above, we propose a method to search for the linear transformation. We first build a comprehensive input set by enumerating a large number of words in the LM's vocabulary. While some words might indicate clear answers for certain knowledge (e.g., \textit{Obama} as $X$ for $F_{\textrm{Country}}$), most of them do not (e.g., \textit{Lit} as $X$ for $F_{\textrm{Country}}$). We feed all inputs to different prompts and collect the output logits such as $\textrm{LogP}_{\textrm{City}}$ and $\textrm{LogP}_{\textrm{Country}}$. For each logit, we only keep dimensions for words falling inside the corresponding output vocabulary domain such as $D_{\textrm{City}}$ and $D_{\textrm{Country}}$. By collecting numerous ($10$K in our experiments) logit pairs, we fit the linearity transformation $(W, b)$ with half of those pairs $(\textrm{LogP}_{\textrm{City, X}}, \textrm{LogP}_{\textrm{Country, X}}), \forall X \in \mbox{Train}$ and then evaluate the transformation on other half of pairs $(\textrm{LogP}_{\textrm{City, X}}, \textrm{LogP}_{\textrm{Country, X}}), \forall X \in \mbox{Test}$.

\paragraph{Evaluation.} With $(W, b)$, we make predictions on the test pairs, $\textrm{LogP}_{\textrm{Country, X}} = W \cdot \textrm{LogP}_{\textrm{City, X}}+b, \forall X \in \mbox{Test}$. We compare the predictions with the test references using the correlation metric, Pearson correlation, to evaluate how similar the logits are distributed. The evaluation is applied by both instance-wise (averaged over instance-wise logits on $x_1, x_2, \cdots \in X$) and label-wise (averaged over label-wise logits across instances on $d_1, d_2, \cdots \in D$). Our main content focuses on the label-wise Pearson correlation as we find that the global bias $b$ plays an important role in the instance-wise predictions as shown in Appendix~\ref{apdx:instance-wise}. The label-wise evaluation eliminates the effect of $b$, which concentrates on the logit correlation matrix $W$. Another advantage of instance-wise correlation is that the metric is calculated based on distributions with the same dimensions. Besides, the correlation weights on different labels also reflects how well each label is approximated by the linear transformation.

\subsection{Experiment Setup}


While numerous compositional knowledge pairs exist in natural language, we focus on large families of knowledge composition that share a commonality. Specifically, we include four large families, attribute, cross-language, simile, and math. We include $111$ prompts in our experiments to cover broad knowledge fields as listed in Appendix~\ref{apdx:prompts}.

\begin{itemize}[nosep,leftmargin=*]
    \item \textbf{Attribute.} Updating one attribute of a subject will affect other attributes as well. The \textit{City$\rightarrow$Country} example illustrated before shows such a compositional relation in the spatial attribute. Another example is shown as follows,
    \small
    $$F_{\textrm{CEO}} \rightarrow F_{\textrm{Company}}$$
    \item \textbf{Cross-language.} Knowledge update is expected to be propagated to other languages, like the $\textit{English}\rightarrow\textit{French}$ example, 
    \small
    $$F_{\textrm{City}} \rightarrow F_{\textrm{Ville}}$$
    \item \textbf{Simile.} Simile builds equivalence among the attributes between objects. Thus, updating a simile to a subject will result in updating the corresponding attribute. An example is as follows,
    \small
    $$F_{\textrm{SameColorAsFruit}} \rightarrow F_{\textrm{Color}}$$
    \item \textbf{Math.} Numbers have denser compositional relations with each other, such as ``\textit{X+1=\underline{2}}''$ \rightarrow$``\textit{X+2=\underline{3}}''. We involve the four basic arithmetic operations in experiments to explore the knowledge composition in math. An example is,
    \small
    $$F_{\textrm{X+1}} \rightarrow F_{\textrm{X+2}}$$
\end{itemize}

\begin{table}
\centering
\small
\scalebox{.95}{
\begin{tabular}{p{1.0cm}p{3.2cm}p{2.5cm}}
\toprule
Family & Prompt & Domain Examples\\
\midrule
\multirow{2}*{Attribute} &  ``X lives in the city of'' & Paris, Vienna\\
 &  ``X lives in the country of'' & France, Austria\\
 \midrule
\multirow{2}*{X-Lang.} &  ``X vit dans la ville de'' & Paris, Vienne\\
 &  ``X lebt in der Stadt von'' & Paris, Wien\\
 \midrule
\multirow{2}*{Simile} &  ``X has the same color as'' & Apple, Banana\\
 &  ``X's color is'' & Red, Yellow\\
 \midrule
\multirow{2}*{Math} &  ``X+1='' & 1, 2, 3, 4, 5\\
 &  ``X*2='' & 2, 4, 6, 8, 10\\
\bottomrule
\end{tabular}
}
\caption{Examples of prompts and domains in different families of knowledge composition.}
\vspace{-5mm}
\label{tab:prompt}
\end{table}

\begin{figure*}
    \centering
    \includegraphics[width=\linewidth]{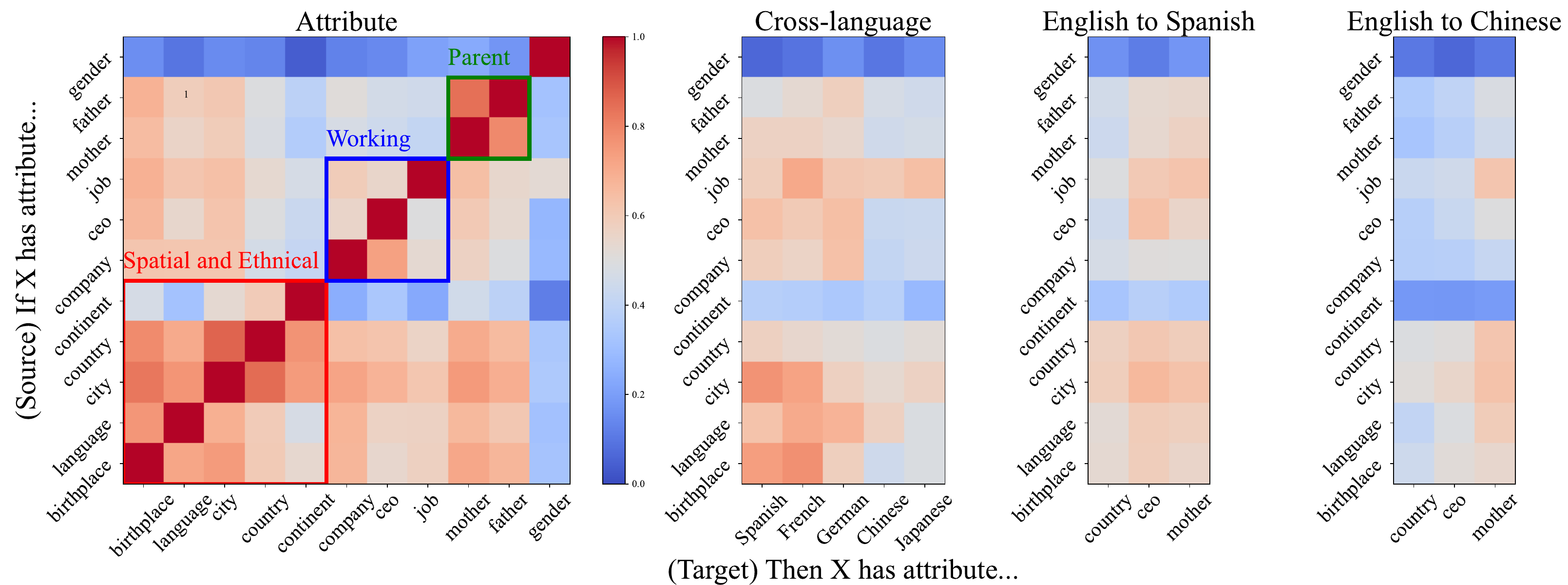}
    \vspace{-7mm}
    \caption{The linear correlation between NTP logits of \texttt{llama-3-8b}.}
    \vspace{-3mm}
    \label{fig:correlation}
\end{figure*}

For each family, we include the results on $10\sim 20$ knowledge prompts in the main content to save the length and place the others in Appendix~\ref{apdx:whole}. Table~\ref{tab:prompt} showcases some examples of prompts and domains.

We include different LLaMA-3~\citep{dubey2024llama3} models in our experiments with parameter numbers of $1$B, $3$B, $8$B, and $70$B. We include the before and after post-training LMs for the evaluation of linear correlation's resilience against fine-tuning. The variance in the model scale allows us to explore the generality and scaling law of the linear correlation inside different models. 
We include LMs from the same family to ensure consistency in tokenization and training data, allowing for a more controlled and convenient discussion. Results on other LMs for broader generality are also included in Appendix~\ref{apdx:whole}.

\subsection{Experiment Results}
\label{subsec:harvest}

The main results of the linear correlation between NTP logits are presented in Figure~\ref{fig:correlation}, where we put a subset of results. The whole massive results are plotted in Figures of Appendix~\ref{apdx:whole}, which are referred to in the main discussions.

\paragraph{Attribute} The correlation pattern between attributes reflects a prominent semantic factor behind the correlation. For instance, the spatial attributes $\{\textit{city}, \textit{country}, \textit{continent}\}$ are highly correlated with each other. Other attributes such as ethical attributes like language also show high correlations with spatial attributes. Besides spatial attributes, there are also highly correlated attribute clusters including job and family-related clusters. On the other hand, we can observe a much weaker correlation between unrelated attributes in the real world, indicating that LMs disentangle the correlation. The gender attribute is a good example, which cannot be identified by any other attribute, showing the effort of the LM to avoid gender bias (except the job attribute as some jobs like \textit{policeman} and \textit{policewoman} can identify the gender). These correlations reflect how knowledge is organized inside the parameters of LMs, which shows high consistency with the real world. There also exist related knowledge with poor correlation like \textit{Language$\rightarrow$Continent} and \textit{CEO$\rightarrow$Company}, reflecting the limitation of LMs in comprehending all knowledge composition.

\paragraph{Cross-language} The results demonstrate some cross-lingual correlation in LMs, which suggests that the knowledge is shared across languages to some degree. However, the correlation between the same concept in different languages is not as strong as related attributes, especially for languages in different families (e.g. \textit{English$\rightarrow$Chinese}). The relatively weak correlation can be attributed to the dominance of English in LLaMA-3 training~\citep{dubey2024llama3}, which we provide further insights using a multilingual LM, Aya~\citep{aya_model}, in Appendix~\ref{apdx:aya}.

\paragraph{Simile} As shown in Table~\ref{tab:correlation_simile} of Appendix~\ref{apdx:main_content}, the correlation between the attribute and the object in the simile also shows a moderate linear correlation. This indicates that LMs bridge an object in similes with its attributes, which is another evidence that LMs can implicitly transfer knowledge. 

\paragraph{Math} The results from the same math operator shows a strong correlation with one another in Figure~\ref{fig:correlation_math} of Appendix~\ref{apdx:main_content}. While this indicates strong mutual influences between calculations, we will show in the next subsection that the correlation in math is imprecise.

\subsection{$W$ can Reflect Real-world Knowledge}
The weight matrix $W$ can reflect compositional relations between source and target domains. Thus, we check whether the $W$'s weights reflect real-world knowledge. Specifically, for each token in the source (target) domain, we check whether the top-influenced (influencing) outputs (inputs), i,e. have the highest weights, are consistent with the real world. We use Hit@Top-$N$ ($N=1,3,5$) metric to evaluate whether there is a correct influenced (influencing) token with a top weight. In experiments that require closed reference, we test subset of knowledge pairs with clear causal relations (e.g., \textit{City}$\rightarrow$\textit{Country} rather than \textit{Mother}$\rightarrow$\textit{Father}). The experiment scale is relatively small due to the sparsity of knowledge composition with clear references.

\begin{table}
\centering
\small
\scalebox{0.92}{
\begin{tabular}{lcccccc}
\toprule
\multirow{4}*{Relation Pair} & \multicolumn{6}{c}{Hit@Top-N} \\
\cmidrule(l){2-7}
 & \multicolumn{3}{c}{Influenced Target} & \multicolumn{3}{c}{Influencing Source} \\
\cmidrule(l){2-4} \cmidrule(l){5-7}
& $1$& $3$& $5$ & $1$& $3$& $5$\\
\midrule
City$\rightarrow$Country & $0.42$ & $0.45$ & $0.48$ & $0.67$ & $0.74$ & $0.78$ \\
CEO$\rightarrow$Company & $0.09$ & $0.09$ & $0.14$ & $0.05$ & $0.05$ & $0.08$ \\
\midrule
City$_{\textrm{en}}$$\rightarrow$City$_{\textrm{es}}$ & $0.91$ & $0.91$ & $0.92$ & $0.67$ & $0.74$ & $0.78$ \\
City$_{\textrm{en}}$$\rightarrow$City$_{\textrm{zh}}$ & $0.10$ & $0.13$ & $0.16$ & $0.09$ & $0.11$ & $0.15$ \\
\midrule
Fruit$\rightarrow$Color & $0.25$ & $0.38$ & $0.47$ & $0.38$ & $0.50$ & $0.54$ \\
Food$\rightarrow$Taste & $0.28$ & $0.50$ & $0.62$ & $0.14$ & $0.36$ & $0.43$ \\
\midrule
X+1$\rightarrow$X+2 & $0.00$ & $0.50$ & $0.60$ & $0.10$ & $0.30$ & $0.50$ \\
X+1$\rightarrow$X*2 & $0.10$ & $0.40$ & $0.50$ & $0.10$ & $0.30$ & $0.70$ \\
\bottomrule
\end{tabular}
}
\caption{The precision of compositional relations built up in $W$.} 
\vspace{-3mm}
\label{tab:w_prec}
\end{table}

We analyze the $W$ precision of $2$ cases from each family with the results presented in Table~\ref{tab:w_prec}. We find the LM have a relatively precise understanding of the correlation between certain highly correlated attributes like \textit{City$\rightarrow$Country}. In transformation matrix $W$, $42\%$ cities learn the top-$1$ weight with their influenced countries and $67\%$ countries have a correct top-$1$ influencing city. For less correlated \textit{CEO$\rightarrow$Company} attributes, $W$ is also imprecise, suggesting the failure to reflect the real-world causal relation. This phenomenon is also observed in the cross-language family for the strongly correlated \textit{English}$\rightarrow$\textit{Spanish} and the weakly correlated \textit{English}$\rightarrow$\textit{Chinese}. However, a strong correlation does not necessarily guarantee a precise $W$ as shown in the math cases.

\begin{table}
\centering
\small
\scalebox{.95}{
\begin{tabular}{cc}
\toprule
If City $=$ & Then Country $=$\\
\midrule
Shanghai & \textcolor{teal}{\textbf{China}},  \textcolor{black}{Italia}, \textcolor{black}{Albania}, \textcolor{black}{USSR}, \textcolor{black}{Korea}\\
NYC & \textcolor{teal}{\textbf{USA}}, \textcolor{black}{USSR}, \textcolor{black}{UAE}, \textcolor{black}{China}, \textcolor{black}{CCCP}\\
Oslo & \textcolor{black}{CCCP}, \textcolor{teal}{\textbf{Norway}}, \textcolor{black}{Kosovo}, \textcolor{black}{Israel}, \textcolor{black}{Oman} \\
Seattle & \textcolor{black}{Uruguay}, \textcolor{black}{Serbia}, \textcolor{black}{Kosovo}, \textcolor{black}{Romania}, \textcolor{black}{Slovenia}\\
Indianapolis & \textcolor{black}{India}, \textcolor{black}{Indonesia}, \textcolor{black}{France}, \textcolor{black}{Iraq}, \textcolor{black}{Netherlands}\\
\midrule
If $X+1=$ & Then $X+2=$\\
\midrule
$1$ & \textcolor{black}{1}, \textcolor{teal}{\textbf{2}}, \textcolor{black}{4}, \textcolor{black}{6}, \textcolor{black}{3}\\
$2$ &  \textcolor{black}{2}, \textcolor{teal}{\textbf{3}}, \textcolor{black}{4}, \textcolor{black}{5}, \textcolor{black}{7} \\
$3$ &  \textcolor{black}{3}, \textcolor{black}{6}, \textcolor{black}{5}, \textcolor{teal}{\textbf{4}}, \textcolor{black}{7}\\
$4$ &  \textcolor{black}{4}, \textcolor{black}{0}, \textcolor{black}{2}, \textcolor{black}{1}, \textcolor{black}{10} \\
$5$ &  \textcolor{black}{5}, \textcolor{teal}{\textbf{6}}, \textcolor{black}{8}, \textcolor{black}{7}, \textcolor{black}{9}\\
\bottomrule
\end{tabular}
}
\caption{Cases of top-influenced tokens pairs in target knowledge.}
\vspace{-3mm}
\label{tab:influence_pair}
\end{table}

In Table~\ref{tab:influence_pair}, we showcase some top-influenced tokens in the attribute and math correlations to visualize how $W$ reflects real-world correlations. In the \textit{City}$\rightarrow$\textit{Country} case, some cities like \textit{Shanghai} and \textit{NYC} are matched with the correct countries while some others like \textit{Oslo}, \textit{Seattle}, and \textit{Indonesia} are not. The \textit{Indonesia$\rightarrow$India} case indicates a bias introduced by superficial similarity into the weights in $W$. The math cases show the correlation is dominated by identical mapping. While the LM tries to model a correct correlation as many secondly influenced numbers are correct, the domination of identical mapping hinders the precision of $W$ to reflect real-world correlation. More cases in Appendix~\ref{apdx:case} further support our observation and extend it to non-causal correlation like parent name correlation.

\subsection{Is $W$ More Accurate in Larger LMs?}
\begin{figure}
    \centering
    \includegraphics[width=\linewidth]{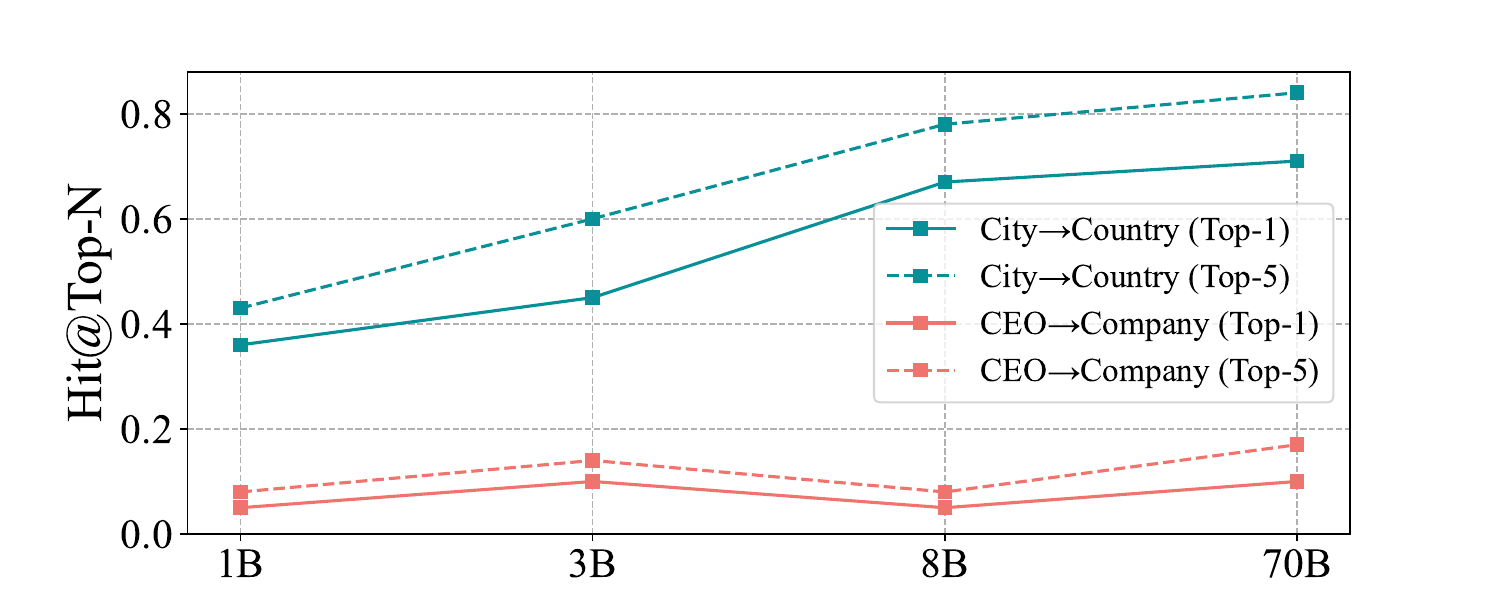}
    \vspace{-5mm}
    \caption{The scaling-up of the precision of $W$ with model size.}
    \label{fig:scale}
\end{figure}

Our discovery indicates that $W$ reflects real-world correlations between knowledge. We check whether the weights of W are more in line with the real world knowledge for larger LMs. Thus, we plot the Top-$N$ metric of correlations in LLaMA-3 of different model sizes in Figure~\ref{fig:scale}. In the \textit{City}$\rightarrow$\textit{Country} case, we can view a clear scaling-up of $W$'s precision, showing that larger LMs also better organize their knowledge. However, \textit{CEO$\rightarrow$Company} is shown to be a hard causal relation, whose $W$'s precision is not successfully scaled up by a larger model size.

\section{Resilient Correlation against Training}
\label{sec:robust}

\subsection{Gradient Correlation}

As many weights in $W$ reflect the real-world correlation, we hypothesize that they are resilient against gradient propagation because they capture inherent patterns that resist change. Thus, we check whether the gradients on related knowledge prompts are also linearly correlated. We choose to train \texttt{llama-3.2-3b}\footnote{We select the smaller 3B LM for fine-tuning efficiency, which shows a similar correlation behavior as the 8B LM in Appendix~\ref{apdx:whole}.} with a common setup for large LMs (AdamW~\citep{AdamW} with $5 \times 10^{-6}$ learning rate). We train the LM with different source knowledge and check whether there is a gradient correlation existing between source and target knowledge. 

\begin{table}
\centering
\small
\scalebox{0.95}{
\begin{tabular}{lcc}
\toprule
Relation Pair & Logit Correlation & Grad. Correlation \\
\midrule
City$\rightarrow$Country & $0.89$ & $0.79$ \\
CEO$\rightarrow$Company & $0.55$ & $0.47$ \\
\midrule
City$_{\textrm{en}}$$\rightarrow$City$_{\textrm{es}}$ & $0.70$ & $0.79$ \\
City$_{\textrm{en}}$$\rightarrow$City$_{\textrm{zh}}$ & $0.58$ & $0.46$ \\
\midrule
Fruit$\rightarrow$Color & $0.48$ & $0.46$ \\
Food$\rightarrow$Taste & $0.47$ & $0.47$ \\
\midrule
X+1$\rightarrow$X+2 & $0.93$ & $0.87$ \\
X+1$\rightarrow$X*2 & $0.73$ & $0.66$ \\
\bottomrule
\end{tabular}
}
\caption{Correlation between gradients on related knowledge.} 
\vspace{-3mm}
\label{tab:grad_corr}
\end{table}

The gradient correlation results are presented in Table~\ref{tab:grad_corr}, demonstrating a correlation between the gradients on different NTP logits. Specifically, with the gradient $\nabla \textrm{LogP}$ on a logit, we can estimate the gradient on a correlated logit by $W\cdot \nabla \textrm{LogP}$. If $W$ is a precise one, the learned knowledge will also be correctly synchronized by knowledge composition caused by $W$, such as Shanghai$\rightarrow$China. Thus, the correlation between gradients indicates a potential mechanism behind how LMs compose learned knowledge. 

\subsection{Correlation after Large-scale Post-training}

We further extend our investigation from a single update to the large-scale post-training of LMs. We check whether the linear correlation is still resilient to large-scale post-training. Thus, we apply the linear transformation $(W, b)$ fitted from an LM before post-training (e.g., \texttt{llama-3-8b}) to its corresponding LM after post-training (e.g., \texttt{llama-3-8b-instruct}). We run the same evaluation as in $\S$~\ref{subsec:harvest} and plot results in Figure~\ref{fig:correlation_crosstuned} of Appendix~\ref{apdx:main_content}. 

Based on the comparison between the two correlation matrices in Figures~\ref{fig:correlation} (before post-training) and ~\ref{fig:correlation_crosstuned} (after post-training), we find the linear transformation still working after numerous optimization steps, indicating $W$ to be resilient against large-scale post-training. This further validates the role of linear correlation in the generalization of LMs, as further discussed in the next section. Another finding in Appendix~\ref{apdx:crosstuned_scale} is that the correlation resilience becomes stronger in larger LMs.

\section{Correlation is a Double-edged Sword}

The potential role of the linear correlation in knowledge composition inspires us to investigate how $W$ implicates the generalization of LMs. We anticipate the resilient correlation to be a two-edged sword, which propagates knowledge with a precise $W$ but also exacerbates hallucination with a imprecise $W$. For validation, we continue to fine-tune the \texttt{llama-3.2-3b} model.


\begin{table}
\centering
\small
\scalebox{0.88}{
\begin{tabular}{cccc}
\toprule
Corr. & Prec. & Relation Pair & Generalization (Random)\\
\midrule
\multirow{3}*{High} & \multirow{3}*{High} & City$\rightarrow$Country & $53.70\%\ (0.78\%)$\\
&  & Country$\rightarrow$Continent & $50.93\%\ (20.00\%)$ \\
&  & City$_{\textrm{en}}$$\rightarrow$City$_{\textrm{es}}$ & $39.10\%\  (0.41\%)$ \\
\midrule
\multirow{2}*{High} &  \multirow{2}*{Low} & X+1$\rightarrow$X+2 & $0.00\%\ (9.09\%)$ \\
 & & X+1$\rightarrow$X*2 & $8.18\%\ (9.09\%)$ \\
 \midrule
\multirow{6}*{Low} & \multirow{6}*{Low} & Fruit$\rightarrow$Color & $11.60\%\ (6.67\%)$ \\
 & & Food$\rightarrow$Taste & $19.44\%\ (10.00\%)$ \\
 & & CEO$\rightarrow$Company & $4.34\%\ (1.00\%)$ \\
 & & Language$\rightarrow$Continent & $23.65\%\ (20.00\%)$ \\
 & & City$_{\textrm{en}}$$\rightarrow$City$_{\textrm{zh}}$ & $2.49\%\ (0.41\%)$ \\
 & & City$_{\textrm{en}}$$\rightarrow$City$_{\textrm{ja}}$ & $4.60\%\ (0.41\%)$ \\
\bottomrule
\end{tabular}
}
\vspace{-2mm}
\caption{The ratio of successful generalization in relation pairs with different linear correlation and $W$ precision.}
\vspace{-5mm}
\label{tab:generalization_ratio}
\end{table}

We first explore how the generalization is affected by the correlation and $W$'s precision. In Table~\ref{tab:generalization_ratio}, we select a relation pair representing high or low in correlation intensity and precision except for the unfounded low correlation and high precision situation. The results show the generalization is only significant when both correlation intensity and $W$'s precision are high. We enumerate more knowledge pairs with low linear correlation than other situation to confirm their poor generalization. This implicates the linear correlation to be an indicator of generalization behavior. When the correlation intensity is high but the $W$'s quality, the LM shows an expectable hallucination. In the \textit{X+1$\rightarrow$X+2} case, learning on any N for ``\textit{X+1=\underline{N}}'' will generalize to a high ``\textit{X+2=\underline{N}}'' as discussed in the case study in Figure~\ref{tab:influence_pair}. 

\begin{figure}
    \centering
    \includegraphics[width=0.9\linewidth]{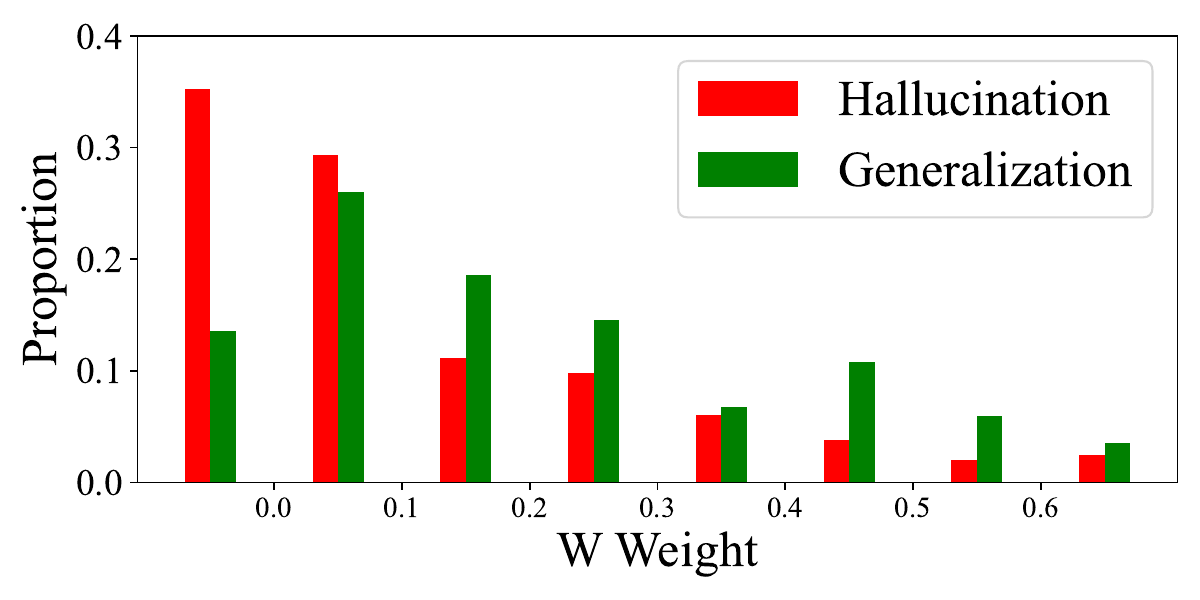}
    \vspace{-5mm}
    \caption{The effect of $W$ weights on generalization.}
    \vspace{-2mm}
    \label{fig:w_generalize}
\end{figure}

For further explanation, we check the weight of ground-truth pairs in the generalized and hallucinated cases of \textit{City$\rightarrow$Country}. As shown in Figure~\ref{fig:w_generalize}, we find the $W$ weight to be an underlying factor in deciding whether the knowledge can be composed. Generally, a higher $W$ weight on the ground-truth pair results in a higher probability to generalization as the gradient will be more efficiently propagated, considering the observed gradient correlation in Table~\ref{tab:grad_corr}.

\begin{table}
\centering
\small
\scalebox{0.9}{
\begin{tabular}{cccccc}
\toprule
City & Reference & Generalized & $W_{\textrm{ref}}$ & $W_{\textrm{gen}}$ & $W_{\textrm{max}}$ \\
\midrule
Shanghai & China & \textcolor{teal}{China} & $0.50$ & $0.50$ & $0.50$ \\
NYC & USA & \textcolor{teal}{USA} & $0.58$ & $0.58$ & $0.58$ \\
Copenhagen & Denmark & \textcolor{teal}{Denmark} & $0.47$ & $0.47$ & $0.47$ \\
Karnataka & India & \textcolor{teal}{India} & $0.34$ & $0.34$ & $0.56$ \\
Indianapolis & USA & \textcolor{red}{India} & $-0.05$ & $0.15$ & $0.17$ \\
Dresden & Germany & \textcolor{red}{Israel} & $0.04$ & $0.13$ & $0.15$ \\
Canberra & Australia & \textcolor{red}{Canada} & $0.04$ & $0.10$ & $0.10$ \\
Helsinki & Finland & \textcolor{red}{Sweden} & $0.42$ & $0.11$ & $0.42$ \\
\bottomrule
\end{tabular}
}
\vspace{-2mm}
\caption{Generalization and hallucination in \textit{City$\rightarrow$Country}.} 
\vspace{-2mm}
\label{tab:case_hallucination}
\end{table}

However, Figure~\ref{fig:w_generalize} also shows that a high $W$ weight does not guarantee a successful generalization. To investigate the underlying reason, we make several case studies in Table~\ref{tab:case_hallucination}. 1) The first $3$ cases illustrate a correct generalization with a top $W$ weight. 2) The fourth case (\textit{Karnataka$\rightarrow$India}) shows a generalization without a top $W$ weight because \textit{India} has a high prior probability (bias) for its high frequency. In contrast, the top-influenced country \textit{Rwanda} has a low prior probability, making the hallucination in the gradient not explicitly propagated into the prediction.

The hallucinated cases can also be divided into two categories. 1) Wrong $W$ weight, a major reason of compositional hallucination. The fifth to seventh cases show low ground-truth $W$ weights, consequently leading to unsuccessful generalization. These cases also show a relatively low maximal weight in $W$, which is potentially an indicator of imprecise $W$ weights. 2) Low prior probability. The last case shows a high $W$ weight between \textit{Helsinki} and \textit{Finland} but the prior probability of \textit{Finland} is much lower than \textit{Sweden}, which results in a compositional hallucination. This is a mirror case of the \textit{Karnataka$\rightarrow$India} generalization.

\section{What Causes the Correlation?}
\label{sec:building}

\begin{figure}
    \centering
    \includegraphics[width=0.9\linewidth]{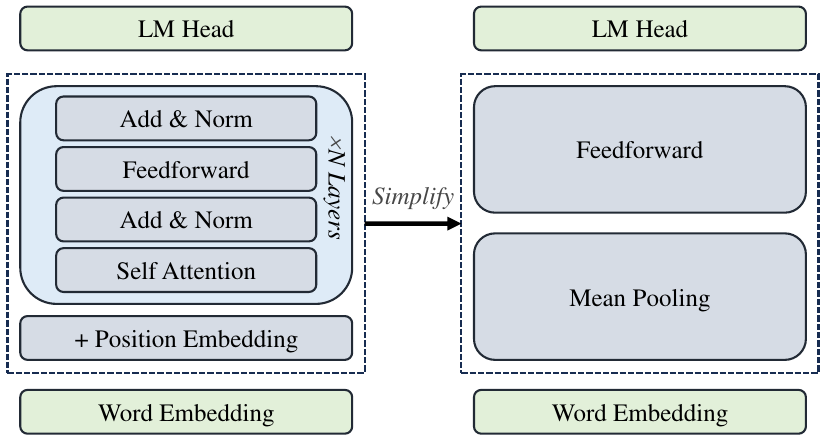}
    \vspace{-3mm}
    \caption{We replace the deep intermediate layers of LMs with an initialized shallow bag-of-word network.}
    \vspace{-2mm}
    \label{fig:ablation_simplify}
\end{figure}

Finally, we investigate the cause behind such linear correlation. Besides the pre-training data distribution, we hypothesize that vocabulary representations play a crucial role in causing such correlations. This is because LMs with different intermediate architectures all show similar correlation behavior in Appendix~\ref{apdx:whole}. To support our hypothesis, we launch a simple ablation study by replacing the complex intermediate architectures (position embedding, self-attention, layer normalization, etc.) of LLaMA-3 with a mean pooling layer and a single initialized feedforward network as shown in the Figure~\ref{fig:ablation_simplify}. To imitate the distribution causing the correlation, the feedforward network is then tuned with $1024$ paired texts such as (``\textit{X lives in the city of \underline{Shanghai}}'', ``\textit{X lives in the city of \underline{China}}'') for $1000$ epochs to learn the knowledge composition relations. For evaluation, the LM is tuned with $128$ source knowledge such as ``\textit{Z lives in the city of \underline{Shanghai}}'' (\textit{Z} different from any $X$ in training) for $2000$ epochs. Then we check whether the LMs can predict composed knowledge, such as ``\textit{Z lives in the city of \underline{China}}''. 

\begin{table}
\centering
\small
\scalebox{0.94}{
\begin{tabular}{lc}
\toprule
Mapping & Generalization \\
\midrule
\textit{(City$\rightarrow$Country)}\\
Shanghai, Tokyo, Paris$\rightarrow$China, Japan, France & $97.66\%$\\
Shanghai, Tokyo, Paris$\rightarrow$Japan, France, China & $22.66\%$\\
S, T, P$\rightarrow$C, J, F & $36.72\%$\\
\midrule
\textit{(Country$\rightarrow$Continent)}\\
China, France, Canada$\rightarrow$Asia, Europe, North & $78.12\%$\\
\midrule
\textit{(CEO$\rightarrow$Company)}\\
Elon, Andy, Tim$\rightarrow$Tesla, Amazon, Apple & $58.59\%$\\
\midrule
\textit{(+1$\rightarrow$+2)}\\
1, 2, 3$\rightarrow$3, 4, 5 & $9.38\%$ \\
\bottomrule
\end{tabular}
}
\vspace{-2mm}
\caption{Generalization with Different Vocabulary Mappings.} 
\vspace{-5mm}
\label{tab:vocabulary_mapping}
\end{table}

Several test results are presented in Table~\ref{tab:vocabulary_mapping}, showing a consistent generalization performance as the initial deep Transformer model. When we switch the correspondence between cities and countries or keep only the first letter, the generalization behavior disappears, which highly attributes the generalization ability to the vocabulary representations.

\section{Conclusion}

This work reveals a new perspective on how LMs generalize by knowledge composition. We detect linear correlations between related NTP logits, which are resilient to training. Such correlations are found to propagate updates on knowledge to one another, leading to compositional generalization and hallucination. We attribute the correlation to vocabulary representations with an ablation study. Future topics include further investigating the formation of such linear correlation and utilizing it for generalizable learning. 

\section*{Impact Statement}

This paper investigates the generalization mechanism behind LMs, which will not explicitly introduce any negative ethical or social impacts. Furthermore, our work have a positive impact on detecting the potential compositional bias caused by unintended correlation with attributes like gender. Fortunately, no current popular LMs show significant compositional bias in gender according to our results.

\nocite{langley00}

\bibliography{example_paper}
\bibliographystyle{icml2024}

\newpage
\appendix
\onecolumn

\section{Results for Main Content}
\label{apdx:main_content}

\begin{figure}
    \centering
    \caption{The linear correlation between NTP logits of \texttt{llama-3-8b} in math operations.}
    \label{fig:correlation_math}
    \includegraphics[width=0.63\linewidth]{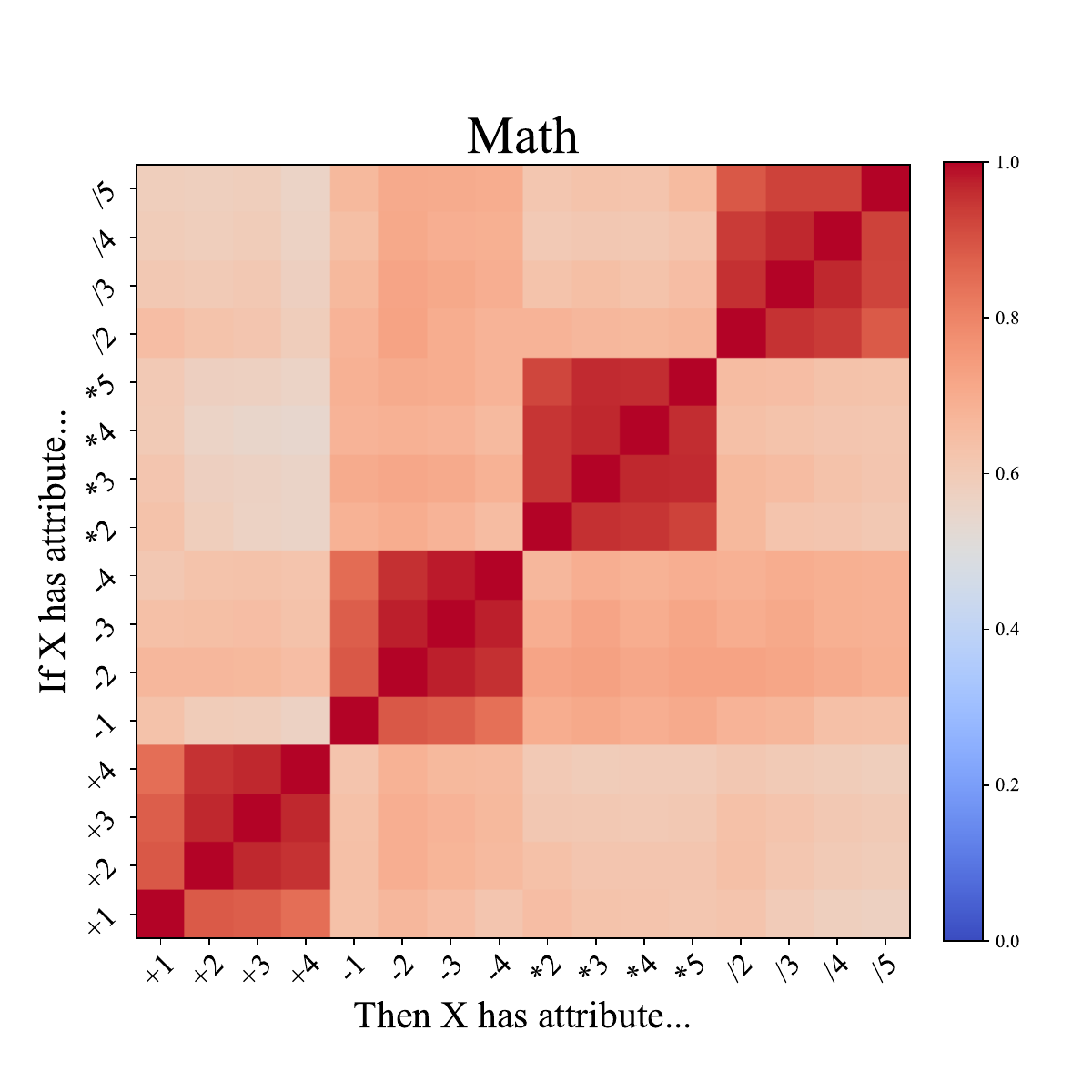}
\end{figure}

\begin{figure}
    \centering
    \caption{The linear correlation between NTP logits of \texttt{llama-3-8b} \textbf{before and after large-scale post-training}.}
    \label{fig:correlation_crosstuned}
    \includegraphics[width=\linewidth]{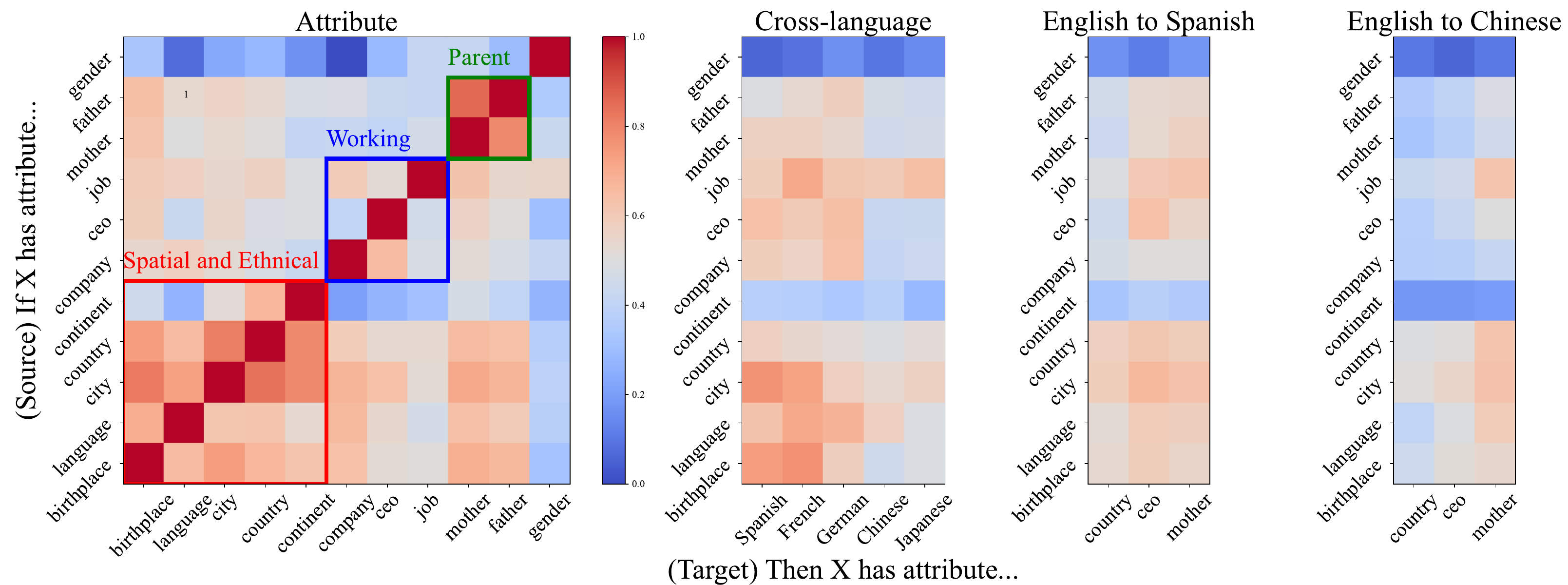}
\end{figure}

\begin{figure}
    \centering
    \caption{The linear correlation between NTP logits in math operations \textbf{before and after large-scale post-training}.}
    \label{fig:correlation_math_crosstuned}
    \includegraphics[width=0.63\linewidth]{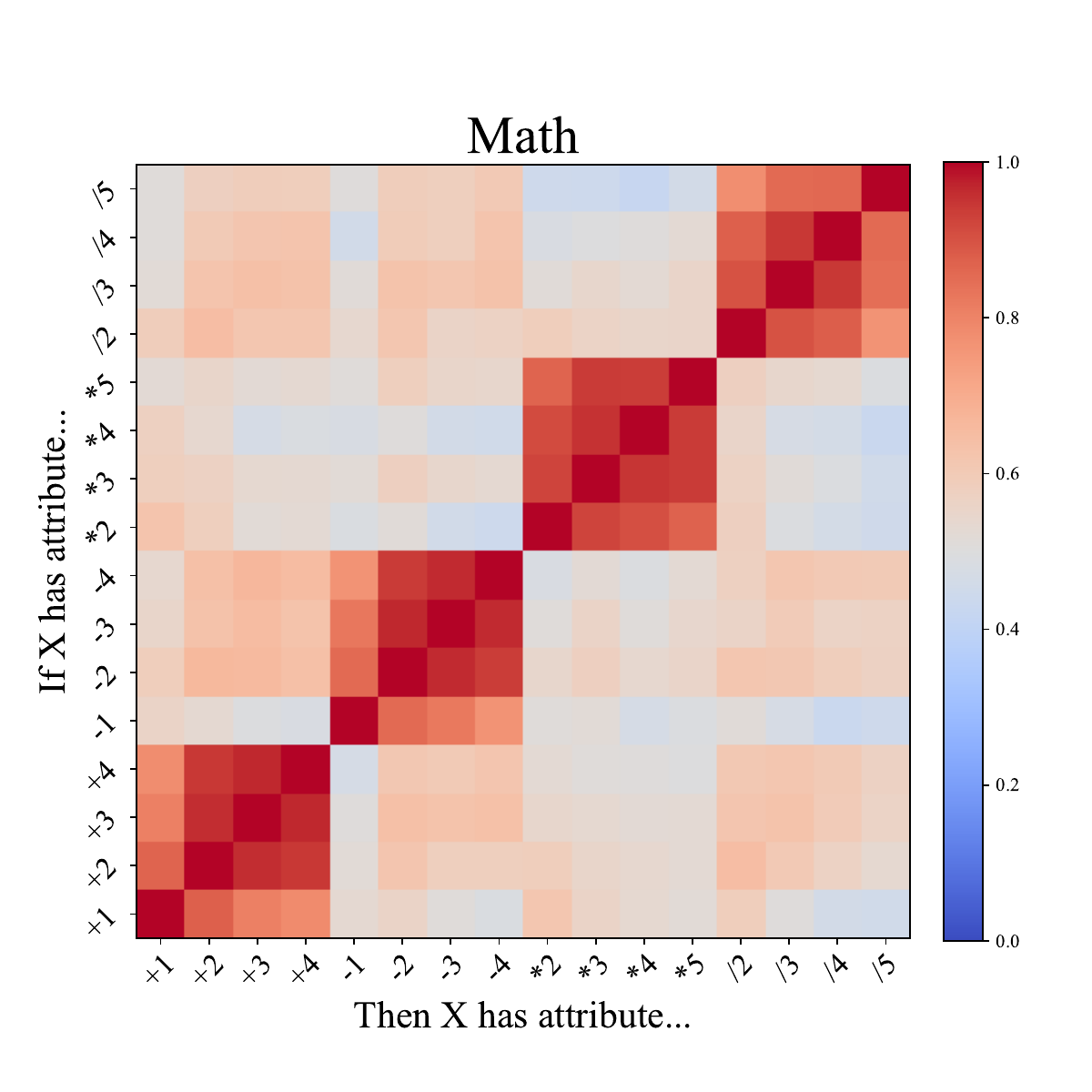}
\end{figure}

In Table~\ref{tab:correlation_simile}, Figures~\ref{fig:correlation_math} and~\ref{fig:correlation_crosstuned}, we illustrate the experiment results for the main content because of the length limitation. Table~\ref{tab:correlation_simile} demonstrates the correlation between simile objects and attributes. Figure~\ref{fig:correlation_math} shows a high correlation between math calculation results. Figure~\ref{fig:correlation_crosstuned} presents the linear correlation between logits from knowledge before and after large-scale post-training, which is compared with the results in Figure~\ref{fig:correlation} to conclude a resilient linear correlation against fine-tuning. The cross-tuning results for simile and math families are presented in Table~\ref{tab:correlation_simile_crosstuned} and Figure~\ref{fig:correlation_math_crosstuned}, which validate a resilient correlation against post-training for highly correlated knowledge pairs. Note that the concepts in \textit{Object} (apple, t-shirt, laptop, chair, washing machine, etc.) for simile relations do not directly indicate attributes, so they are not used for evaluation when reference is required.

\begin{table}
\centering
\small
\caption{Correlation between gradients on simile objects and attributes.} 
\label{tab:correlation_simile}
\scalebox{1.0}{
\begin{tabular}{lccccc}
\toprule
Relation Pair & Fruit-Color & Food-Taste & Gem-Color & Name-Country & Animal-Size \\
\midrule
Correlation & $48.42$ & $46.68$ & $27.46$ & $67.35$ & $59.59$ \\
\midrule
\midrule
Relation Pair & Object-Genre & Object-Heat & Object-Size & Object-Price & Object-Color \\
\midrule
Correlation & $77.68$ & $73.11$ & $71.41$ & $72.87$ & $70.87$ \\
\bottomrule
\end{tabular}
}
\end{table}

\newpage

\begin{table}
\centering
\small
\caption{Correlation between logits on simile objects and attributes \textbf{before and after large-scale post-training}.} 
\label{tab:correlation_simile_crosstuned}
\scalebox{1.0}{
\begin{tabular}{lccccc}
\toprule
Relation Pair & Fruit-Color & Food-Taste & Gem-Color & Name-Country & Animal-Size \\
\midrule
Correlation & $44.11$ & $37.06$ & $33.66$ & $67.30$ & $49.65$ \\
\midrule
\midrule
Relation Pair & Object-Genre & Object-Heat & Object-Size & Object-Price & Object-Color \\
\midrule
Correlation & $72.03$ & $63.75$ & $66.13$ & $71.09$ & $66.27$ \\
\bottomrule
\end{tabular}
}
\end{table}

\clearpage

\section{Limitation and Future Works}

As a pioneering study, our work focuses on uncovering the phenomenon of linear correlations in language models but leaves several key aspects for future research:

\begin{itemize}[nosep,leftmargin=*]
    \item \textbf{Theoretical Explanation} We do not provide a formal theory explaining why resilient linear correlations emerge. Future work can explore the underlying model architectures, optimization dynamics, and linguistic structures that drive this phenomenon.
    \item \textbf{Data Distribution Effects} Our study does not systematically analyze how training data influences the formation of these correlations. Investigating which data properties contribute to their emergence could provide deeper insights.
    \item \textbf{Identifying Correlated Knowledge Pairs} While we observe linear correlations in specific cases (e.g., city–country), we do not establish a general method to predict what knowledge pairs exhibit this property. Future work can develop theoretical or empirical criteria for identifying such relationships.
\end{itemize}

Due to content limitations, we focus on describing the phenomenon rather than fully explaining its origins. We hope our findings serve as a foundation for further research into the mechanisms and implications of linear correlations in LMs.

\section{Prompts and Setups}
\label{apdx:prompts}

\begin{table}
\centering
\small
\scalebox{1.0}{
\begin{tabular}{lc}
\toprule
Template & Domain Size \\
\midrule
Attribute & $23$ \\
Cross-language & $11\times 5 = 55$ \\
Simile & $17$ \\
Math & $4\times4=16$ \\
\midrule
Total & $111$ \\
\bottomrule
\end{tabular}
}
\caption{The statistics of prompts in different families.}
\label{tab:stats}
\end{table}

\begin{table}
\centering
\small
\scalebox{1.0}{
\begin{tabular}{lccc}
\toprule
\multicolumn{2}{c}{Knowledge} & Template & Domain Size \\
\midrule
\multirow{23}*{\rotatebox{90}{\textrm{Attribute}}}& birthplace & ``\textit{\{\} was born in the city of}'' & 242 \\
& city & ``\textit{\{\} lives in the city of}'' & 242 \\
& country & ``\textit{\{\} lives in the country of}'' & 128 \\
& continent & ``\textit{\{\} lives in the continent of}'' & 6 \\
& language & ``\textit{\{\} speaks the language of}'' & 217 \\
& company & ``\textit{\{\} works for the company of}'' & 100 \\
& landmark & ``\textit{\{\} lives near the landmark of}'' & 100 \\
& ceo & ``\textit{\{\} works for the CEO called}'' & 101 \\
& mother & ``\textit{\{\}'s mother's name is}'' & 100 \\
& father & ``\textit{\{\}'s father's name is}'' & 100 \\
& job & ``\textit{\{\}'s job is}'' & 105 \\
& personality & ``\textit{\{\}'s personality is}'' & 100 \\
& pet & ``\textit{\{\}'s pet is}'' & 100 \\
& sport & ``\textit{\{\}'s favorite sport is}'' & 102 \\
& food & ``\textit{\{\}'s favorite food is}'' & 104 \\
& drink & ``\textit{\{\}'s favorite drink is}'' & 102 \\
& gender & ``\textit{\{\}'s gender is}'' & 3 \\
& vehicle & ``\textit{\{\}'s preferred mode of transportation is}'' & 51 \\
& color & ``\textit{\{\}'s favorite color is}'' & 15 \\
& music & ``\textit{\{\}'s favorite music genre is}'' & 100 \\
& hobby & ``\textit{\{\}'s favorite hobby is}'' & 101 \\
& flower & ``\textit{\{\}'s favorite flower is}'' & 97 \\
& vacation & ``\textit{\{\}'s favorite vacation spot is}'' & 101 \\
\bottomrule
\end{tabular}
}
\caption{Templates used in our experiments (Part 1: Attribute).}
\label{tab:prompt_p1}
\end{table}

\begin{table}
\centering
\small
\scalebox{1.0}{
\begin{tabular}{lccc}
\toprule
\multicolumn{2}{c}{Knowledge} & Template & Domain Size \\
\midrule
\multirow{11}*{\rotatebox{90}{\textrm{Spanish}}}& birthplace & ``\textit{\{\} nació en la ciudad de}'' & 242 \\
& city & ``\textit{\{\} vive en la ciudad de}'' & 242 \\
& country & ``\textit{\{\} vive en el país de}'' & 128 \\
& continent & ``\textit{\{\} vive en el continente de}'' & 6 \\
& language & ``\textit{\{\} habla el idioma de}'' & 217 \\
& company & ``\textit{\{\} trabaja para la empresa de}'' & 100 \\
& ceo & ``\textit{\{\} trabaja para el CEO llamado}'' & 101 \\
& job & ``\textit{El trabajo de \{\} es}'' & 105 \\
& mother & ``\textit{El nombre de la madre de \{\} es}'' & 100 \\
& father & ``\textit{\{\} el nombre del padre es}'' & 100 \\
& gender & ``\textit{El género de \{\} es}'' & 3 \\
\midrule
\multirow{11}*{\rotatebox{90}{\textrm{French}}} & birthplace & ``\textit{\{\} est né dans la ville de}'' & 242 \\
& city & ``\textit{\{\} vit dans la ville de}'' & 242 \\
& country & ``\textit{\{\} vit dans le pays de}'' & 128 \\
& continent & ``\textit{\{\} vit sur le continent de}'' & 6 \\
& language & ``\textit{\{\} parle la langue de}'' & 217 \\
& company & ``\textit{\{\} travaille pour l'entreprise de}'' & 100 \\
& ceo & ``\textit{\{\} travaille pour le PDG appelé}'' & 101 \\
& job & ``\textit{\{\} travaille comme}'' & 105 \\
& mother & ``\textit{Le nom de la mère de \{\} est}'' & 100 \\
& father & ``\textit{Le nom du père de \{\} est}'' & 100 \\
& gender & ``\textit{\{\} est de sexe}'' & 3 \\
\midrule
\multirow{11}*{\rotatebox{90}{\textrm{German}}} & birthplace & ``\textit{\{\} wurde in der Stadt geboren}'' & 242 \\
& city & ``\textit{\{\} lebt in der Stadt}'' & 242 \\
& country & ``\textit{\{\} lebt im Land}'' & 128 \\
& continent & ``\textit{\{\} lebt auf dem Kontinent}'' & 6 \\
& language & ``\textit{\{\} spricht die Sprache von}'' & 217 \\
& company & ``\textit{\{\} arbeitet für das Unternehmen von}'' & 100 \\
& ceo & ``\textit{\{\} arbeitet für den CEO namens}'' & 101 \\
& job & ``\textit{Der Beruf von \{\} ist}'' & 105 \\
& mother & ``\textit{Der Name von \{\}'s Mutter ist}'' & 100 \\
& father & ``\textit{Der Name von \{\}'s Vater ist}'' & 100 \\
& gender & ``\textit{Das Geschlecht von \{\} ist}'' & 3 \\
\midrule
\multirow{11}*{\rotatebox{90}{\textrm{Chinese}}} & birthplace & \begin{CJK}{UTF8}{gbsn}``\textit{\{\}所出生的城市是}''\end{CJK} & 242 \\
& city & \begin{CJK}{UTF8}{gbsn}``\textit{\{\}所居住的城市是}''\end{CJK} & 242 \\
& country & \begin{CJK}{UTF8}{gbsn}``\textit{\{\}所居住的国家是}''\end{CJK} & 128 \\
& continent & \begin{CJK}{UTF8}{gbsn}``\textit{\{\}所居住的大陆是}''\end{CJK} & 6 \\
& language & \begin{CJK}{UTF8}{gbsn}``\textit{\{\}说的语言是}''\end{CJK} & 217 \\
& company & \begin{CJK}{UTF8}{gbsn}``\textit{\{\}工作的公司是}''\end{CJK} & 100 \\
& ceo & \begin{CJK}{UTF8}{gbsn}``\textit{\{\}工作的公司的CEO是}''\end{CJK} & 101 \\
& job & \begin{CJK}{UTF8}{gbsn}``\textit{\{\}的工作是}''\end{CJK} & 105 \\
& mother & \begin{CJK}{UTF8}{gbsn}``\textit{\{\}的母亲的名字是}''\end{CJK} & 100 \\
& father & \begin{CJK}{UTF8}{gbsn}``\textit{\{\}的父亲的名字是}''\end{CJK} & 100 \\
& gender & \begin{CJK}{UTF8}{gbsn}``\textit{\{\}的性别是}''\end{CJK} & 3 \\
\midrule
\multirow{11}*{\rotatebox{90}{\textrm{Japanese}}} & birthplace & \begin{CJK}{UTF8}{min}``\textit{\{\}が生まれた都市は}''\end{CJK} & 242 \\
& city & \begin{CJK}{UTF8}{min}``\textit{\{\}が住んでいる都市は}''\end{CJK} & 242 \\
& country & \begin{CJK}{UTF8}{min}``\textit{\{\}が住んでいる国は}''\end{CJK} & 128 \\
& continent & \begin{CJK}{UTF8}{min}``\textit{\{\}が住んでいる大陸は}''\end{CJK} & 6 \\
& language & \begin{CJK}{UTF8}{min}``\textit{\{\}が話している言語は}''\end{CJK} & 217 \\
& company & \begin{CJK}{UTF8}{min}``\textit{\{\}が働いている会社は}''\end{CJK} & 100 \\
& ceo & \begin{CJK}{UTF8}{min}``\textit{\{\}が働いている会社のCEOは}''\end{CJK} & 101 \\
& job & \begin{CJK}{UTF8}{min}``\textit{\{\}の仕事は}''\end{CJK} & 105 \\
& mother & \begin{CJK}{UTF8}{min}``\textit{\{\}の母の名前は}''\end{CJK} & 100 \\
& father & \begin{CJK}{UTF8}{min}``\textit{\{\}の父の名前は}''\end{CJK} & 100 \\
& gender & \begin{CJK}{UTF8}{min}``\textit{\{\}の性別は}''\end{CJK} & 3 \\
\bottomrule
\end{tabular}
}
\caption{Templates used in our experiments (Part 2: Cross Language).}
\label{tab:prompt_p2}
\end{table}

\begin{table}
\centering
\small
\scalebox{1.0}{
\begin{tabular}{lccc}
\toprule
\multicolumn{2}{c}{Knowledge} & Template & Domain Size \\
\midrule
\multirow{17}*{\rotatebox{90}{\textrm{Simile}}} & object\_color & ``\textit{The color of \{\} is the same as}'' & 85 \\
& object\_price & ``\textit{The size of \{\} is the same as}'' & 85 \\
& object\_heat & ``\textit{The heat of \{\} is the same as}'' & 85 \\
& object\_genre & ``\textit{The genre of \{\} is the same as}'' & 85 \\
& object\_size & ``\textit{The size of \{\} is the same as}'' & 85 \\
& simile\_color & ``\textit{The color of \{\} is}'' & 15 \\
& simile\_price & ``\textit{The size of \{\} is}'' & 2 \\
& simile\_heat & ``\textit{The heat of \{\} is}'' & 4 \\
& simile\_genre & ``\textit{The genre of \{\} is}'' & 22 \\
& simile\_size & ``\textit{The size of \{\} is}'' & 3 \\
& simile\_taste & ``\textit{The taste of \{\} is}'' & 3 \\
& name\_country & ``\textit{\{\} lives in the same country as}'' & 128 \\
& gem\_color & ``\textit{The color of \{\} is the same as the gem called}'' & $50$ \\
& animal\_size & ``\textit{The size of \{\} is the same as the animal called}'' & $100$ \\
& food\_taste & ``\textit{\{\} has the same taste as the food:}'' & $95$ \\
& fruit\_color & ``\textit{\{\} X has the same color as the fruit:}'' & $99$ \\
\midrule
\multirow{4}*{\rotatebox{90}{\textrm{Math}}} & X+N & ``\textit{\{\}+N=}'' & 11 \\
 & X-N & ``\textit{\{\}-N=}'' & 11 \\
 & X*N & ``\textit{\{\}*N=}'' & 11 \\
 & X/N & ``\textit{\{\}/N=}'' & 11 \\
\bottomrule
\end{tabular}
}
\caption{Templates used in our experiments (Part 3: Simile and Math).}
\label{tab:prompt_p3}
\end{table}

Table~\ref{tab:stats} shows the statistics of the prompts used in our experiments. Tables~\ref{tab:prompt_p1}, ~\ref{tab:prompt_p2}, ~\ref{tab:prompt_p3} further list all the specific prompts used in our experiments. The domain size of most prompts is around $100$ expect for some domains with limited valid outputs like \textit{Continent} and \textit{Color}.

\newpage

\section{Instance-wise Correlation}
\label{apdx:instance-wise}

\begin{figure*}
    \centering
    \includegraphics[width=0.91\linewidth]{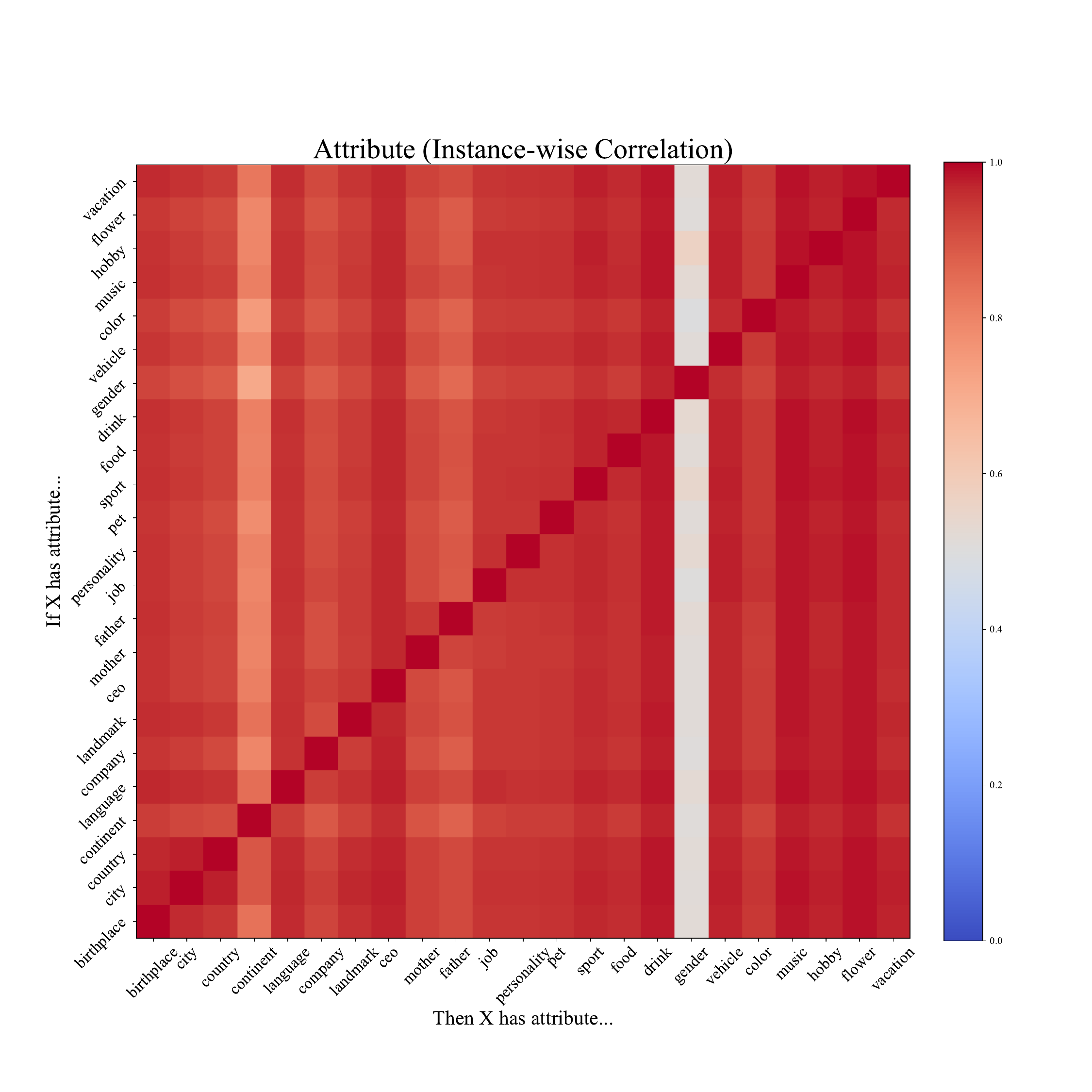}
    \vspace{-5mm}
    \caption{The instance-wise correlation between NTP logits of \texttt{llama3-8b} (attribute 
 as an example).}
    \label{fig:correlation_instance}
\end{figure*}

Figure~\ref{fig:correlation_instance} shows the instance-wise Pearson correlation evaluation results on different knowledge pairs. We use attribute correlation as an example to show that the target knowledge of each instance can be well approximated by a linear transformation on the source knowledge. In the main content, we demonstrate the label-wise correlation because we find the bias term $b$ to dominate the prediction on many knowledge pairs that are poorly linear correlated (especially in gradient). Some target knowledge is predictable with only the prior probability from bias even without any linear indicator. Thus, the label-wise correlation is a more challenging metric by eliminating the effect of $b$ with a better reflection of how the source knowledge influences the target knowledge.

\newpage

\section{Subdomain Building Procedure}
\label{apdx:subdomain}

To build the subdomains, we do not simply collect the top predictions from the next token predictions because many predictions are introduced by the frequency and similarity bias (e.g., stop words like \textit{the}) in the next token representation space~\citep{demeter2020stolen,corr_navi}. Instead, we enumerate the common answers by \texttt{gpt-4o}~\citep{achiam2023gpt4} and search engines. Then we keep the first tokens of the tokenization for these answers which are not subwords. For example,\textit{China} will be represented by \textit{China}, \textit{South Korean} will be represented by \textit{South}, and \textit{Brunei} will be dropped because it is tokenized into [\textit{Br}, \textit{unei}]. We exclude subwords because they cannot identify complete semantics without tokens after them. The discussion for subword cases is included in Appendix~\ref{apdx:subword}.

\section{Whole Attribute Results and Extra Discussion}
\label{apdx:whole}

\begin{figure*}
    \centering
    \includegraphics[width=0.81\linewidth]{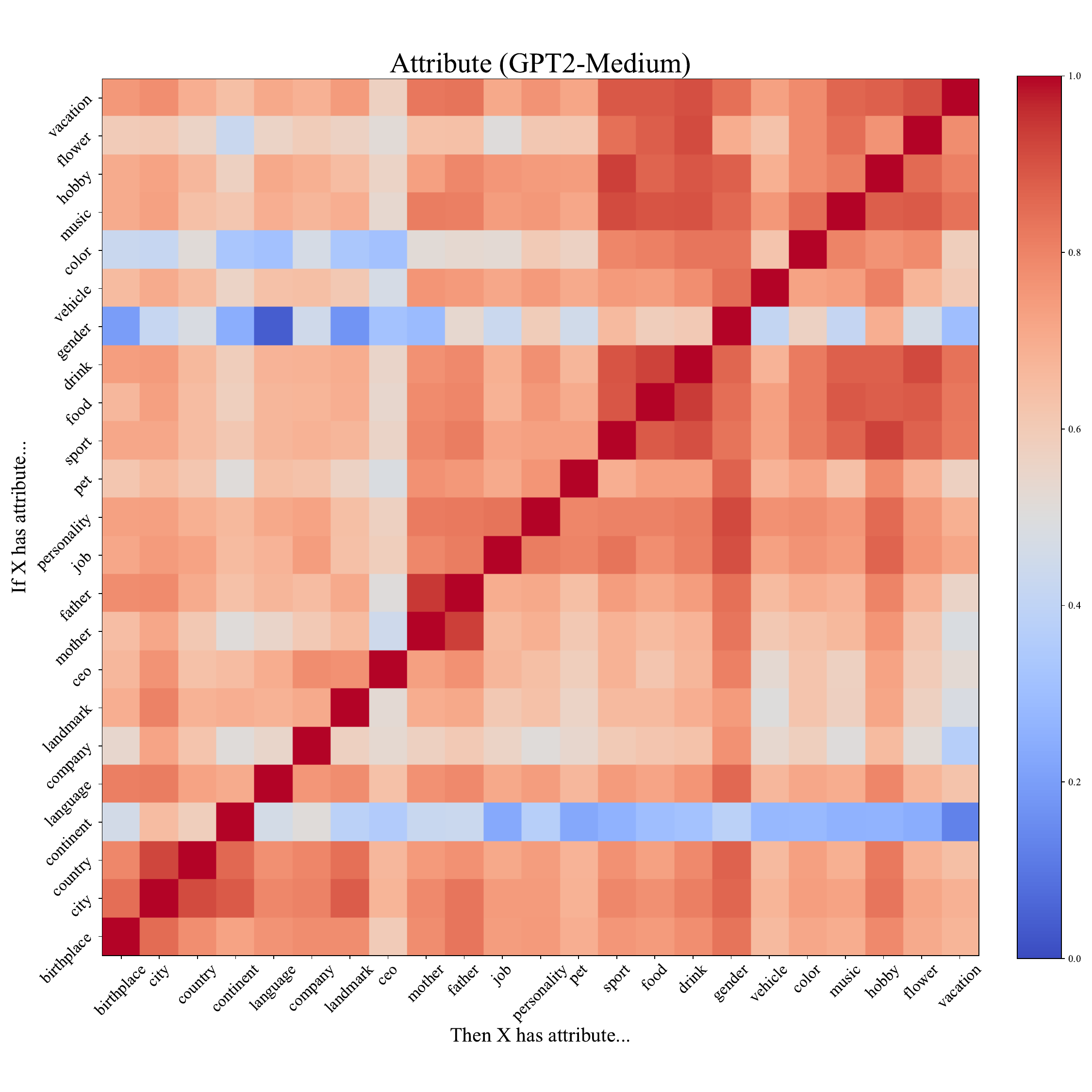}
    \vspace{-5mm}
    \caption{The attribute correlation between NTP logits of \texttt{gpt2-medium}.}
    \label{fig:correlation_gpt2m}
\end{figure*}

\begin{figure*}
    \centering
    \includegraphics[width=\linewidth]{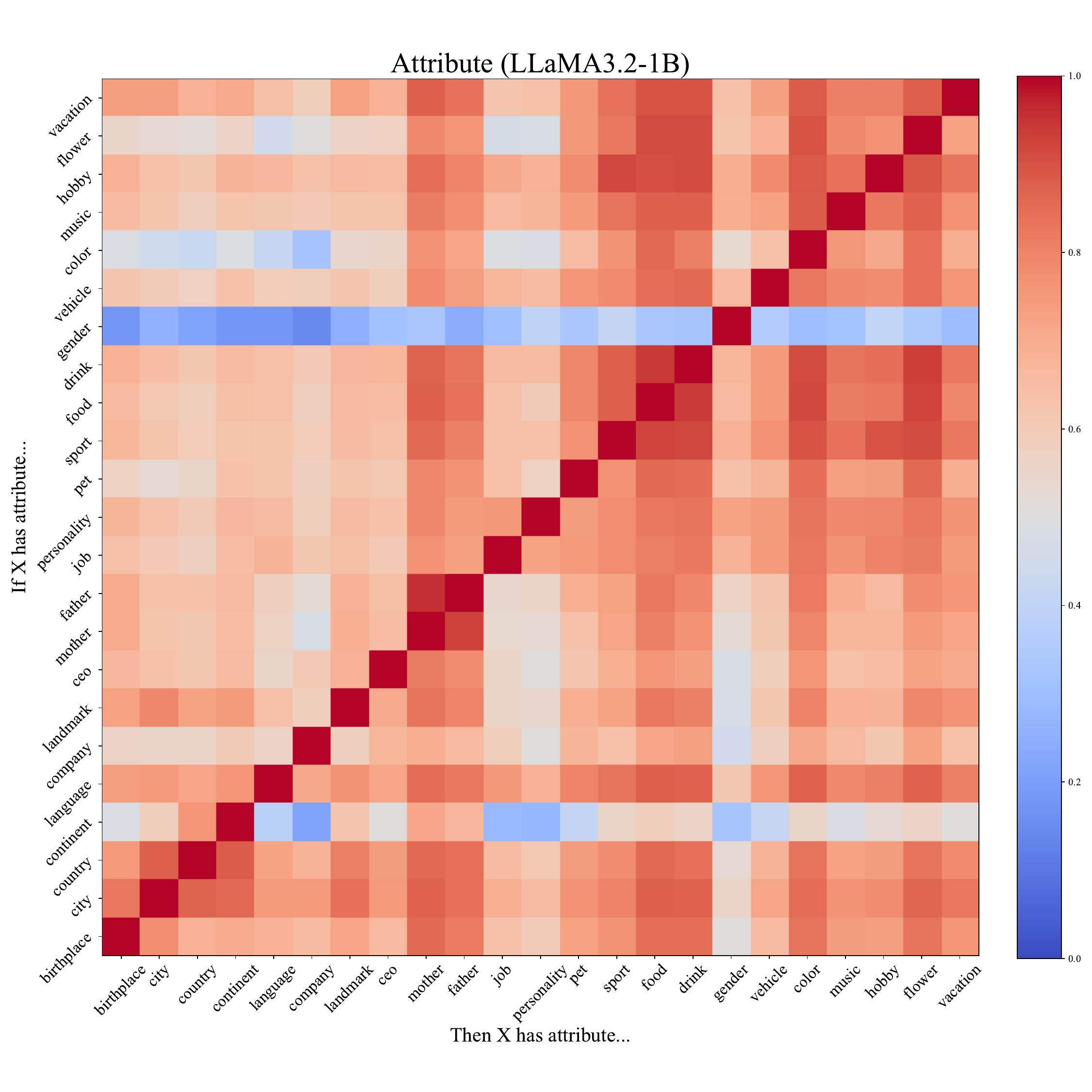}
    \vspace{-5mm}
    \caption{The attribute correlation between NTP logits of \texttt{llama-3.2-1b}.}
    \label{fig:correlation_1b}
\end{figure*}

\begin{figure*}
    \centering
    \includegraphics[width=\linewidth]{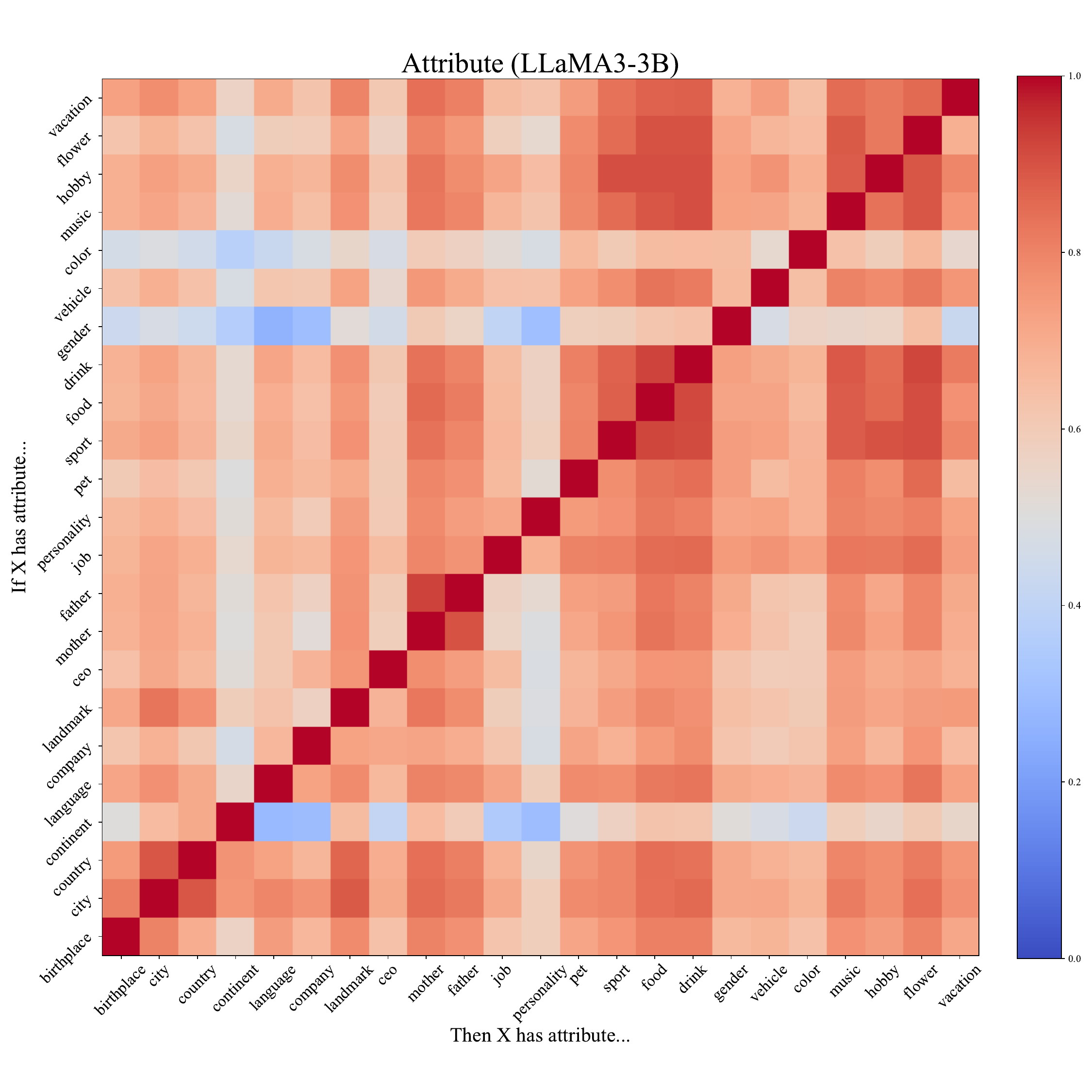}
    \vspace{-5mm}
    \caption{The attribute correlation between NTP logits of \texttt{llama-3.2-3b}.}
    \label{fig:correlation_3b}
\end{figure*}

\begin{figure*}
    \centering
    \includegraphics[width=\linewidth]{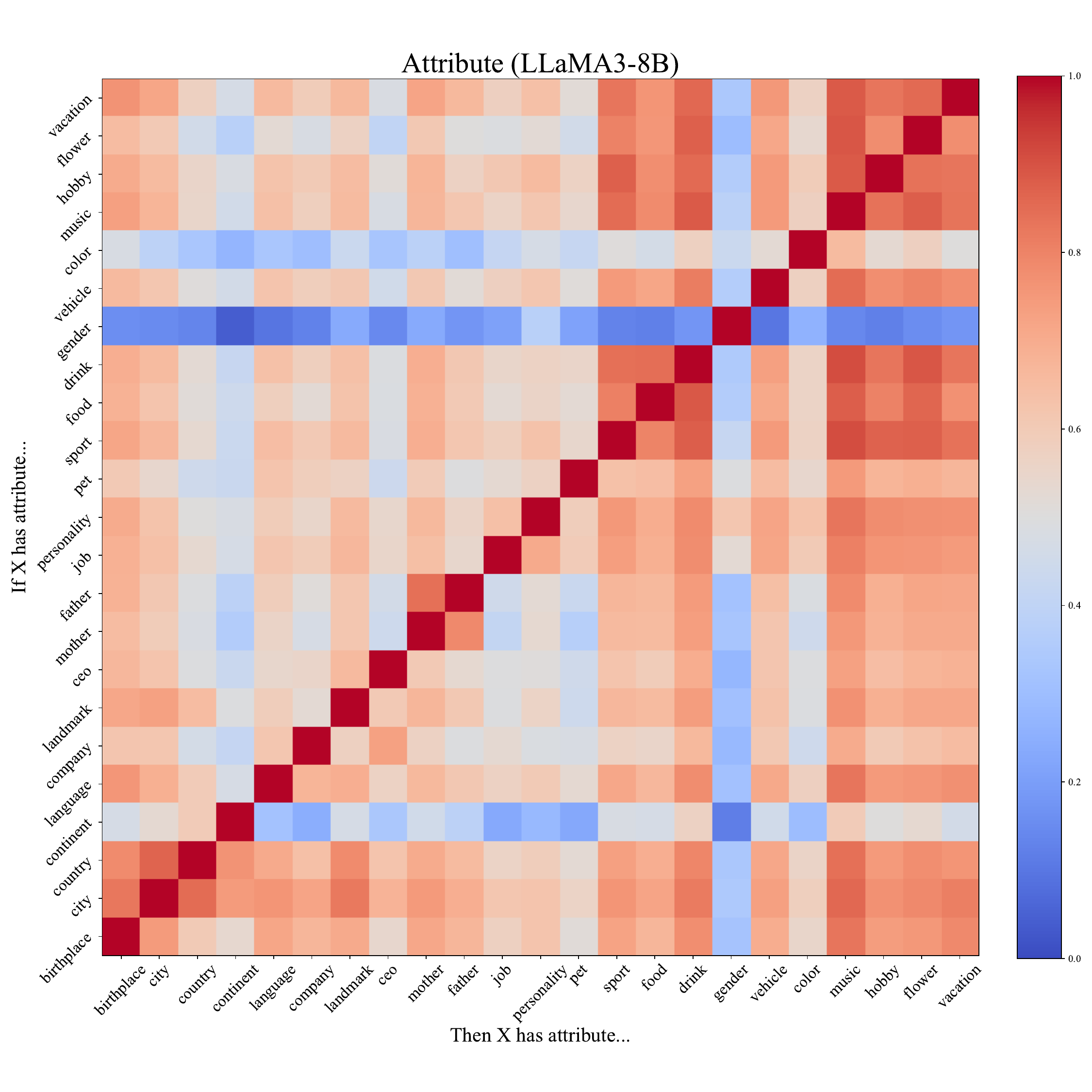}
    \vspace{-5mm}
    \caption{The attribute correlation between NTP logits of \texttt{llama-3-8b}.}
    \label{fig:correlation_8b}
\end{figure*}

\begin{figure*}
    \centering
    \includegraphics[width=\linewidth]{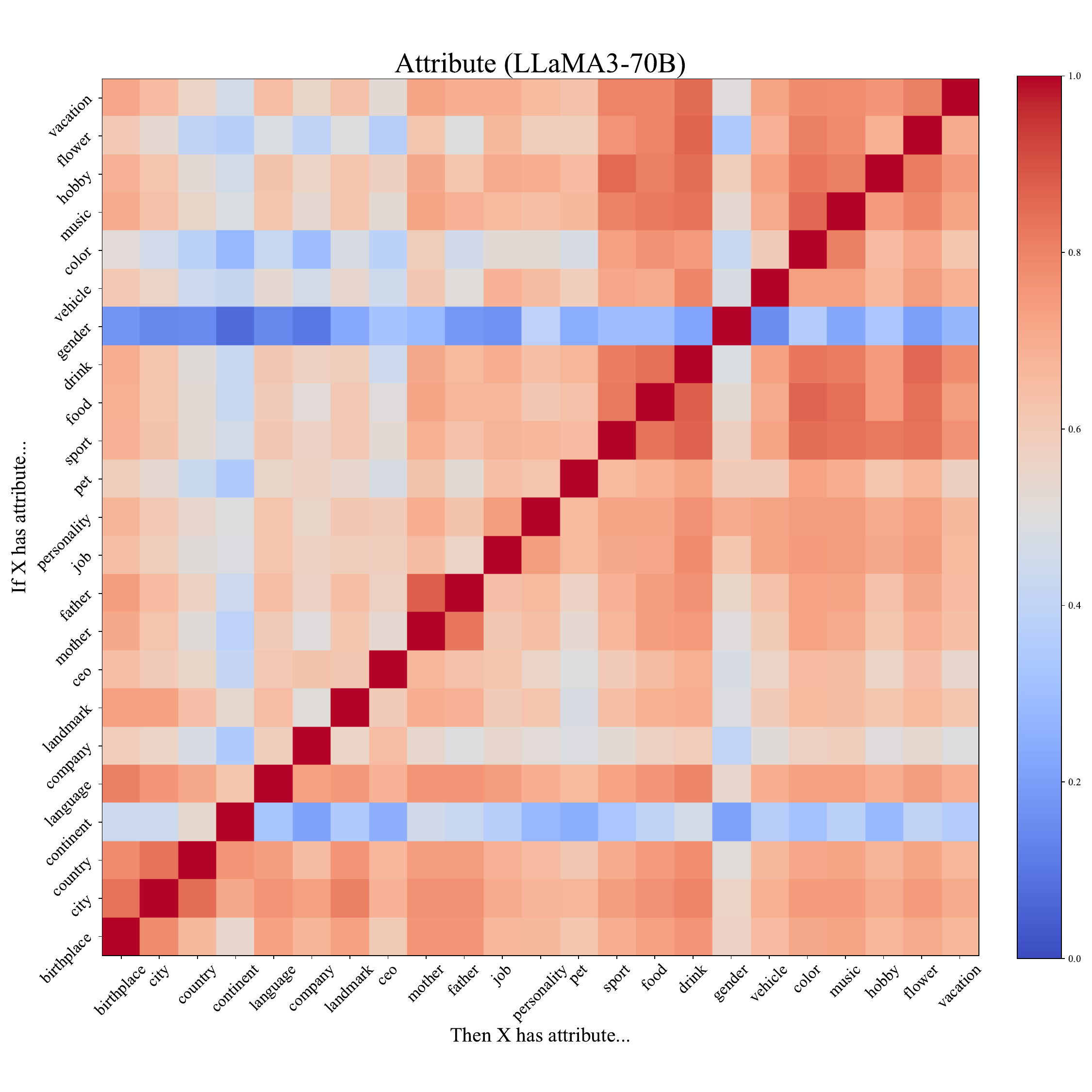}
    \vspace{-5mm}
    \caption{The attribute correlation between NTP logits of \texttt{llama-3-70b}.}
    \label{fig:correlation_70b}
\end{figure*}

\begin{figure*}
    \centering
    \includegraphics[width=\linewidth]{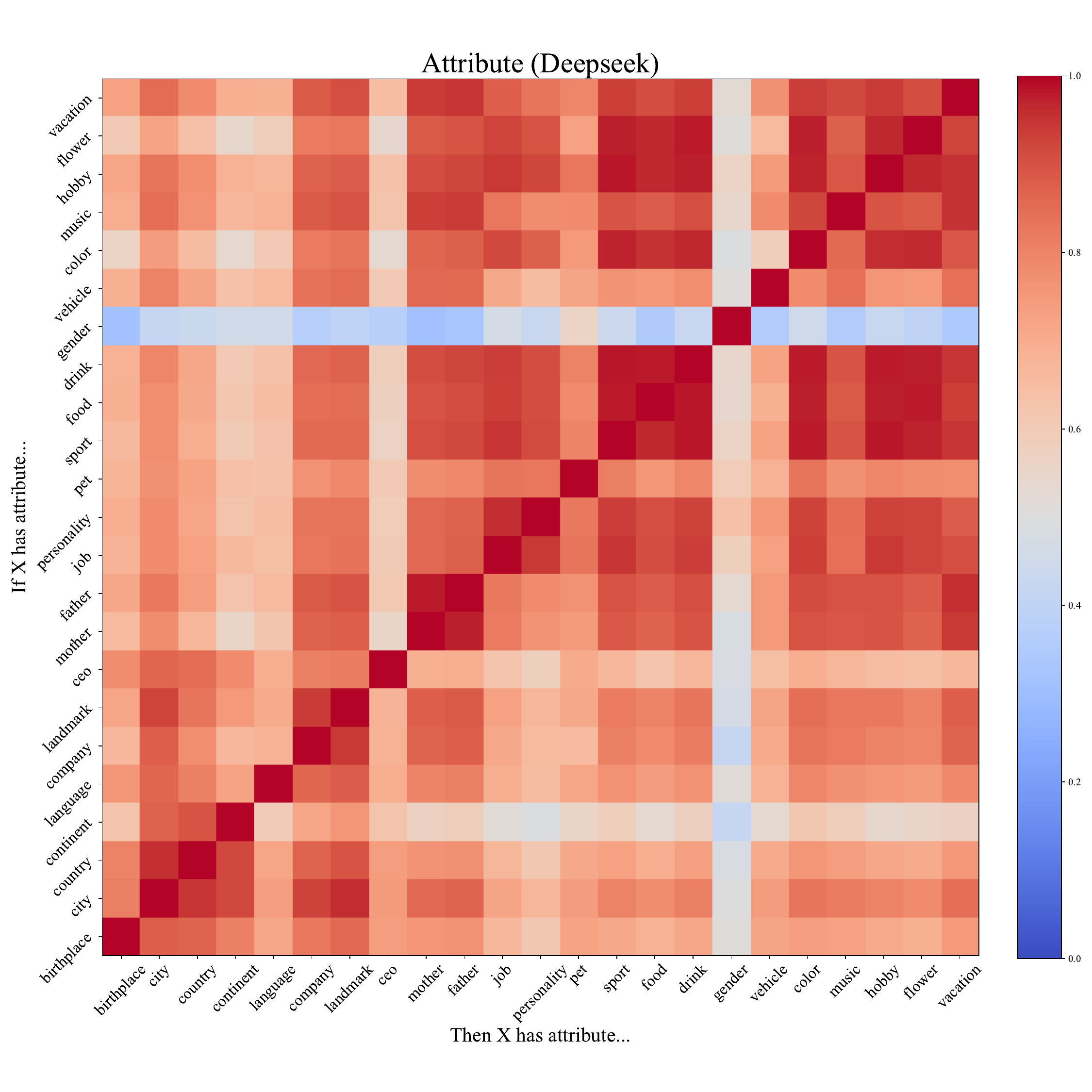}
    \vspace{-5mm}
    \caption{The attribute correlation between NTP logits of \texttt{deepseek-r1-distll-qwen-7B}.}
    \label{fig:correlation_deepseek}
\end{figure*}

\begin{figure*}
    \centering
    \includegraphics[width=\linewidth]{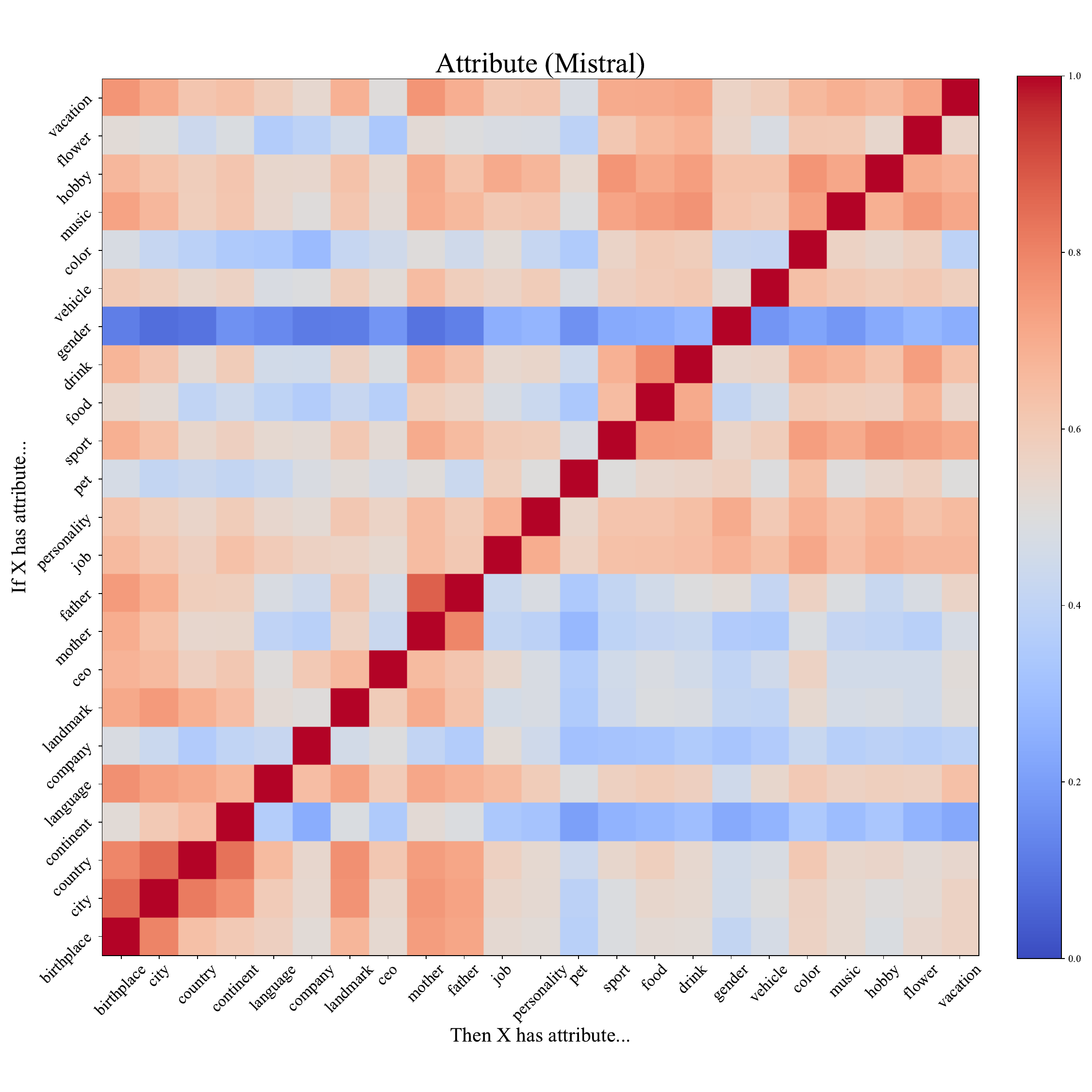}
    \vspace{-5mm}
    \caption{The attribute correlation between NTP logits of \texttt{mistral-7b-v0.3}.}
    \label{fig:correlation_mistral}
\end{figure*}

\begin{figure}
    \centering
    \includegraphics[width=\linewidth]{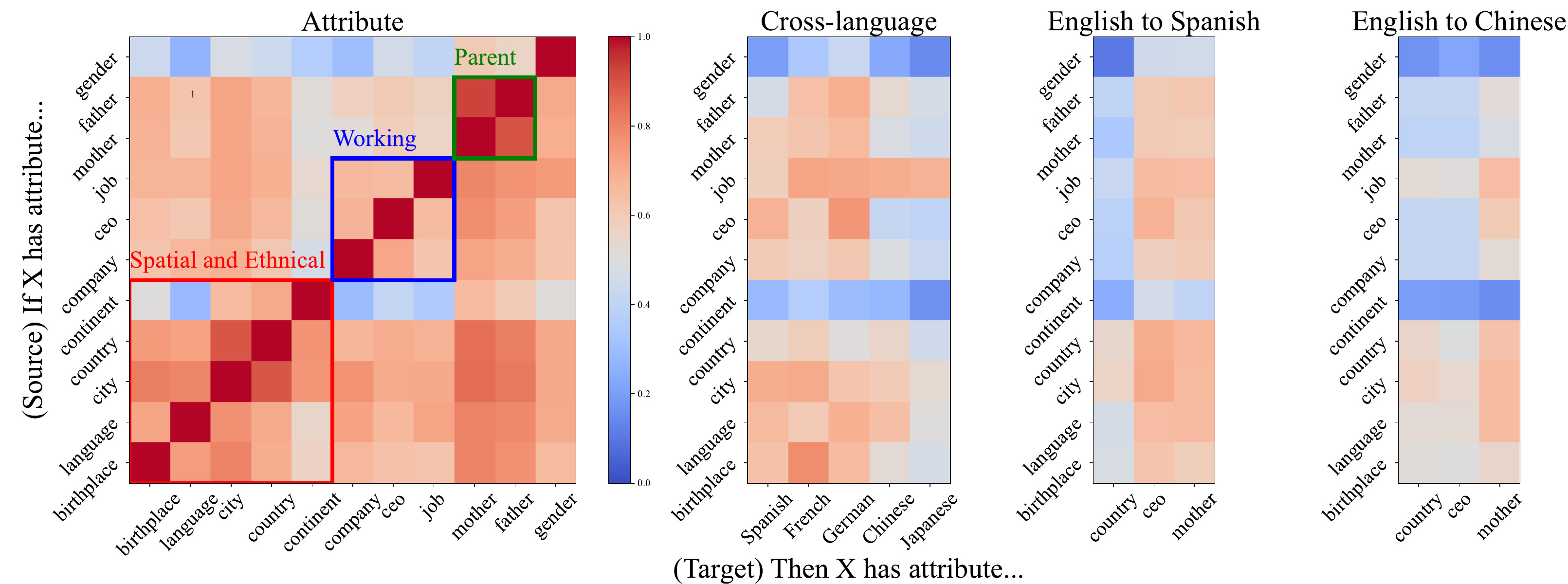}
    \caption{The linear correlation between NTP logits of \texttt{llama-3.2-3b}.}
    \label{fig:correlation_main_3b}
\end{figure}

\begin{figure}
    \centering
    \includegraphics[width=0.8\linewidth]{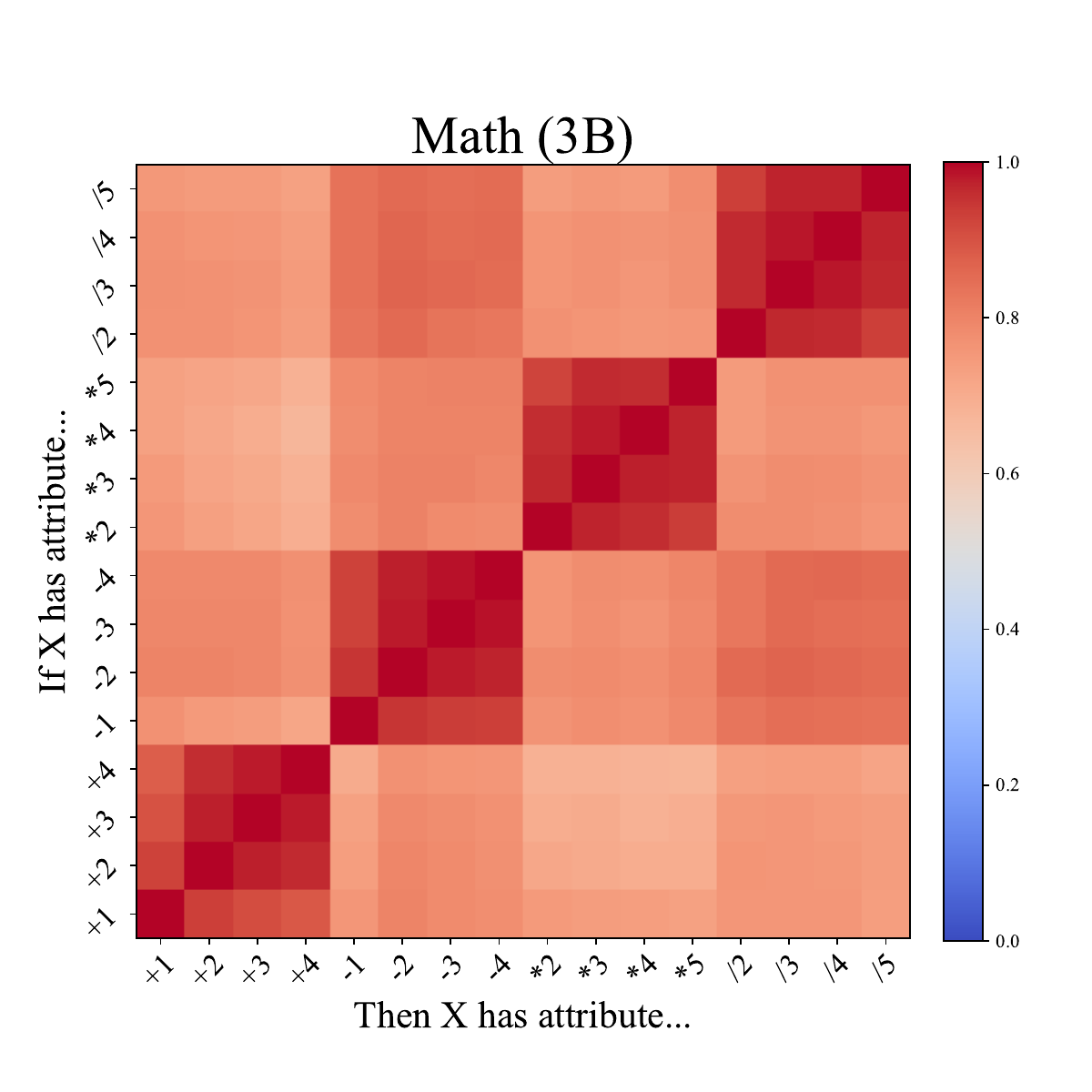}
    \caption{The linear correlation between NTP logits of \texttt{llama-3.2-3b} before and after large-scale post-training.}
    \label{fig:correlation_math_3b}
\end{figure}

From Figure~\ref{fig:correlation_gpt2m} to Figure~\ref{fig:correlation_math_3b}, we present the whole correlation matrices inside all kinds of LMs for different prompts. We can observe the existence of correlation behavior among different LMs. While the correlation in different LMs behaves differently, some common pairs like \textit{City}$\rightarrow$\textit{Country} hold for all different LMs. Also, models from the same LLaMA-3 family tend to behave in a similar way. We can also observe many spurious correlations such as \textit{Hobby}$\rightarrow$\textit{Mother}, which generally have low causal relations in the real world. Larger LMs tend to be better at disentangling such kind of spurious correlations as the smallest \texttt{GPT2-Medium} model shows a much stronger correlation. In Figures~\ref{fig:correlation_main_3b} and~\ref{fig:correlation_math_3b}, Table~\ref{tab:correlation_simile_3b}, we illustrate that the 3B model has a similar correlation behavior as the 8B one. 

\clearpage

\begin{table}
\centering
\small
\scalebox{1.0}{
\begin{tabular}{lccccc}
\toprule
Relation Pair & Fruit-Color & Food-Taste & Gem-Color & Name-Country & Animal-Size \\
\midrule
Correlation & $48.37$ & $46.95$ & $50.48$ & $78.83$ & $69.43$ \\
\midrule
\midrule
Relation Pair & Object-Genre & Object-Heat & Object-Size & Object-Price & Object-Color \\
\midrule
Correlation & $81.92$ & $76.48$ & $84.23$ & $84.23$ & $81.08$ \\
\bottomrule
\end{tabular}
}
\caption{Correlation between logits of \texttt{llama-3.2-3b} on simile objects and attributes.} 
\label{tab:correlation_simile_3b}
\end{table}

\clearpage

\section{More Resilient Correlation in Larger LMs}
\label{apdx:crosstuned_scale}

\begin{figure}
    \centering
    \caption{The correlation becomes more resilient in larger LMs.}
    \label{fig:crosstuned_scale}
    \includegraphics[width=0.48\linewidth]{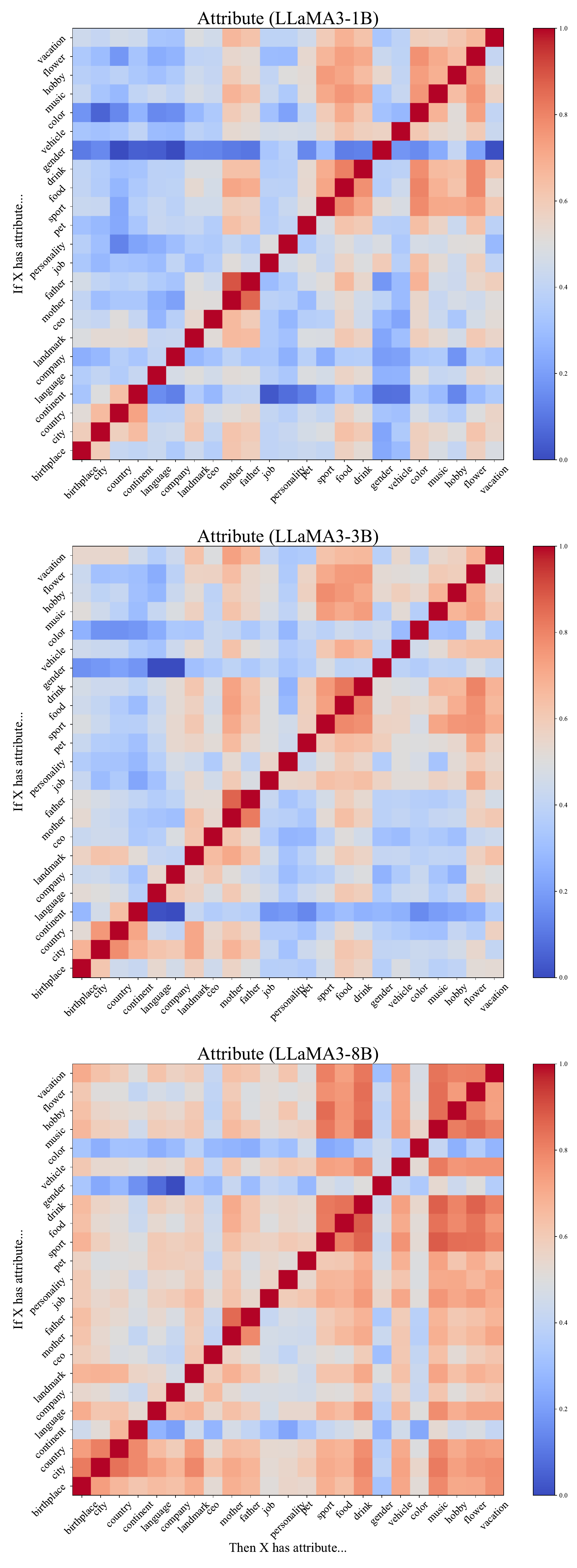}
    \vspace{-5mm}
\end{figure}

\begin{figure}
    \centering
    \caption{The correlation between logits from \texttt{mistral-7b-v0.3} before and after post-training.}
    \label{fig:crosstuned_mistral}
    \includegraphics[width=\linewidth]{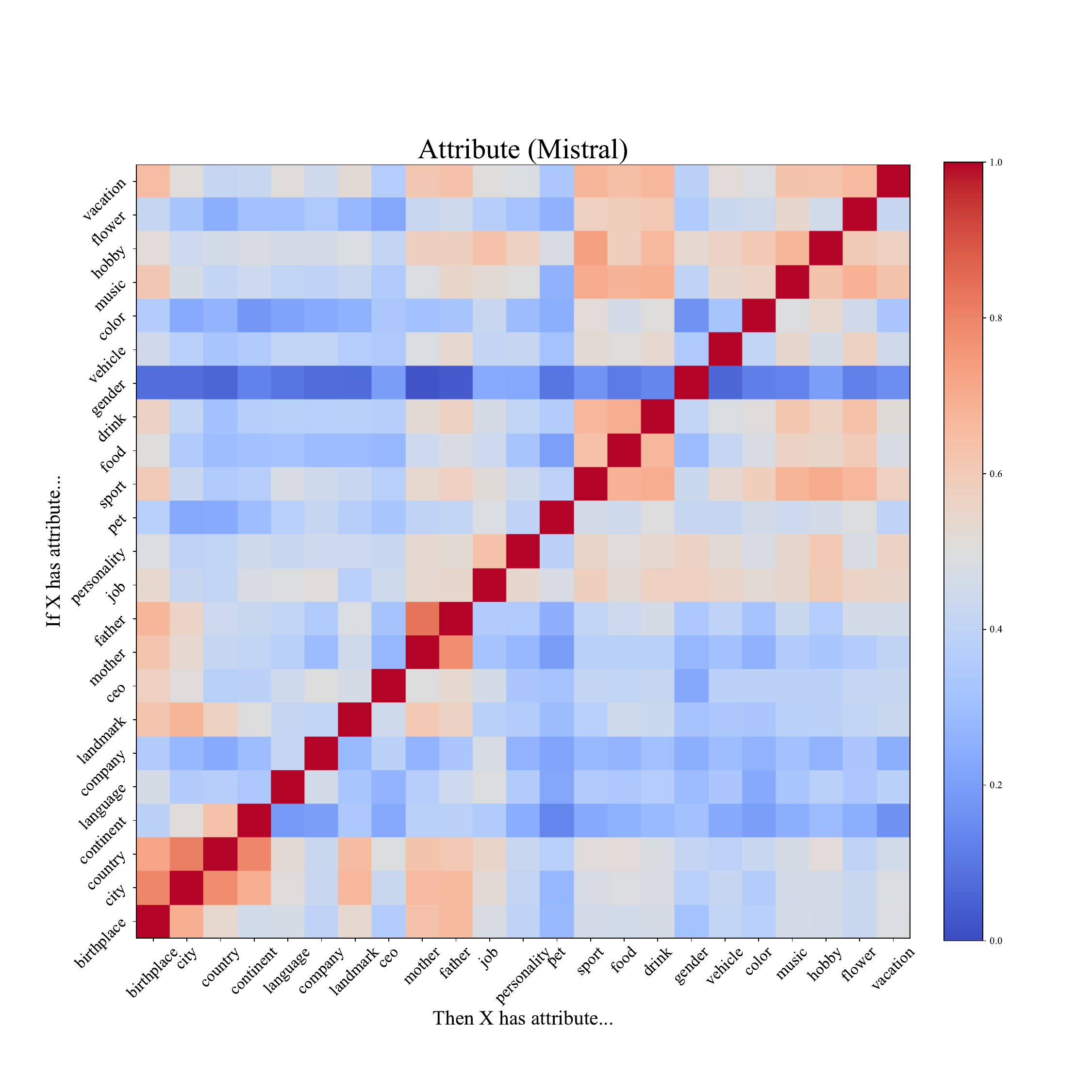}
    \vspace{-5mm}
\end{figure}

In Figure~\ref{fig:crosstuned_scale}, we find the linear correlation is more resilient against fine-tuning by plotting the correlation before and after post-training in $1$B, $3$B, $8$B LLaMA-3 LMs as we find more strong correlations in larger LMs. In Figure~\ref{fig:crosstuned_mistral}, we also plot the correlation matrix between logits from \texttt{mistral-7b-v0.3} before and after post-training, which supports the existence of resilient linear correlation in LMs with vocabulary representation untied.

\clearpage

\section{Multilingual LM}
\label{apdx:aya}

\begin{figure*}
    \centering
    \includegraphics[width=0.5\linewidth]{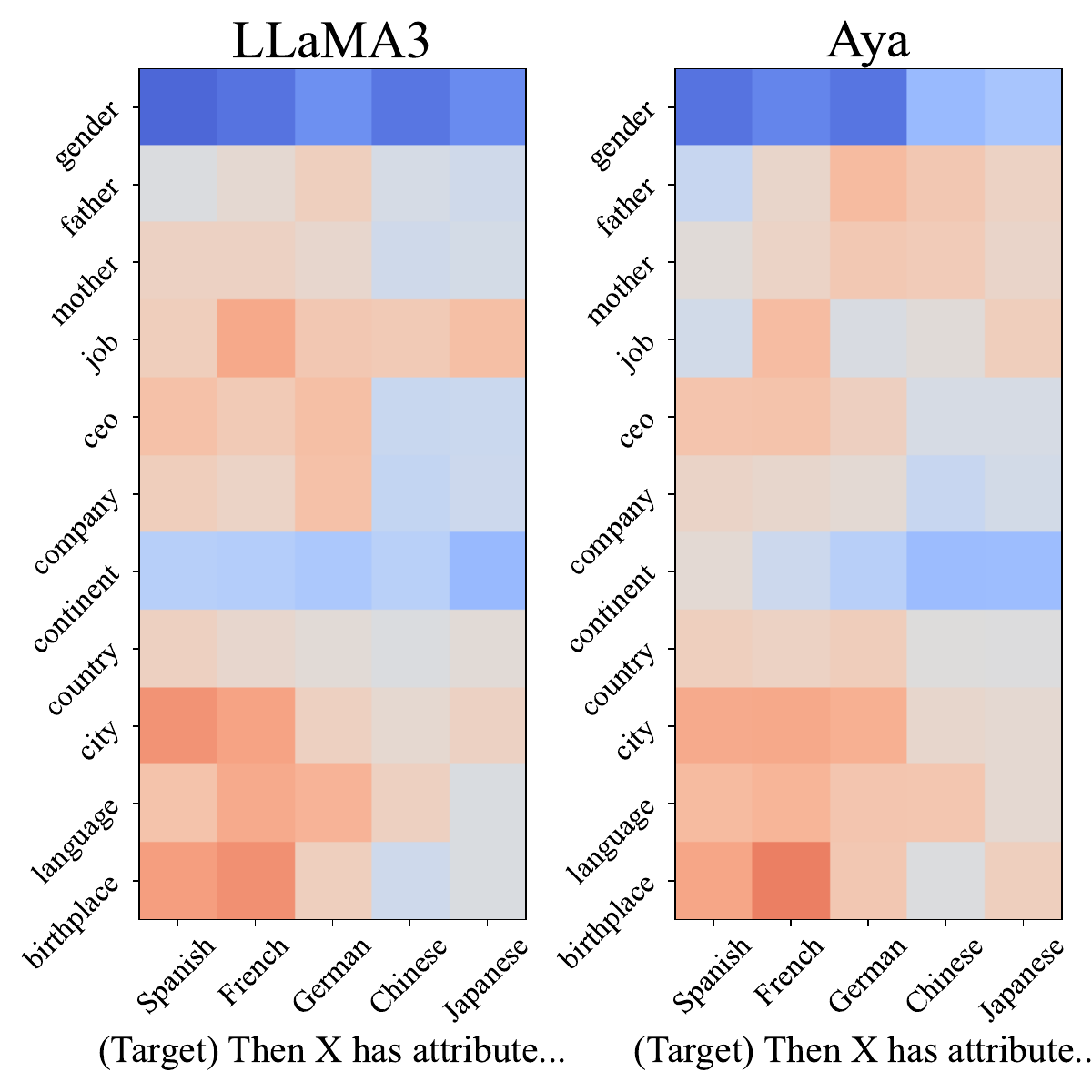}
    \vspace{-5mm}
    \caption{The comparison between Aya and LLaMA in cross-lingual correlation.}
    \label{fig:correlation_aya}
\end{figure*}

Figure~\ref{apdx:aya} demonstrates the cross-lingual correlation of the multilingual LM, \texttt{aya-expanse-8b}, which outperforms LLaMA-3 in multilingual tasks but still lags behind in English~\citep{aya_model}. The results show Aya to have a stronger cross-lingual correlation between knowledge pairs, especially in Chinese and Japanese. On Latin language, Aya's advantage becomes smaller because these languages share quite a lot entity names with English and LLaMA-3 can benefit from its English ability to complement the weakness in multi-lingual ability. 

\clearpage

\section{Extra Case Study}
\label{apdx:case}

We provide extra cases for analysis in this section. In Table~\ref{tab:country_city_case}, we provide massive cases on the influencing cities in the \textit{City$\rightarrow$Country} knowledge composition, which shows that the LM establishes correlation between many (City, Country) pairs such as (\textit{Edinburgh}, \textit{Scotland}), (\textit{Islamabad}, \textit{Pakistan}), and (\textit{Afghanistan}, \textit{Kabul}). Tables~\ref{tab:parent_case} and~\ref{tab:simile_case} showcase the correlation between knowledge pairs that do not have a clear reference. Taking parent correlation as an example, Table~\ref{tab:parent_case} shows correlation of parent names from the same ethnicity like (\textit{Chen}, \textit{Mei}) and (\textit{Santiago}, \textit{Sofia}).

\begin{table}
\centering
\small
\scalebox{0.8}{
\begin{tabular}{cc}
\toprule
Country & Influencing Cities \\
\midrule
 Sweden &  \textcolor{teal}{\textbf{Stockholm}}, Brisbane, Johannesburg, Cardiff, Chicago, Hyderabad, Aleppo, Lima, Rochester, Salem \\
 Cuba &  \textcolor{teal}{\textbf{Havana}}, Chicago, Columbus, stockholm, Rochester, Hyderabad, Scarborough, Johannesburg, singapore, Hamburg \\
 Switzerland &  Columbus, Stuttgart, Cardiff, Leicester, Chicago, Brisbane, Saras, stockholm, vegas, Bethlehem \\
 Ghana &  Winnipeg, Nairobi, Johannesburg, Leicester, Atlanta, Tulsa, Maharashtra, Greenville, Brisbane, Lima \\
 Poland &  \textcolor{teal}{\textbf{Warsaw}}, Cardiff, Liverpool, Maharashtra, stockholm, Amsterdam, Atlanta, Kashmir, Perth, Aleppo \\
 Turkey &  \textcolor{teal}{\textbf{Istanbul}}, Chicago, Toronto, Maharashtra, stockholm, Johannesburg, Cardiff, Lima, Columbus, Ankara \\
 Sudan &  Nairobi, stockholm, Lima, Tulsa, Johannesburg, Maharashtra, Winnipeg, Hyderabad, Wilmington, Kashmir \\
 Romania &  Cardiff, Rochester, Johannesburg, \textcolor{teal}{\textbf{Budapest}}, Seattle, Rajasthan, Hyderabad, Chicago, Kyoto, Lima \\
 Samoa &  Maharashtra, Leicester, Winnipeg, Chicago, Honolulu, Brisbane, Nairobi, Hyderabad, Lima, Cardiff \\
 Iceland &  Cardiff, Leicester, Chicago, Amsterdam, Wilmington, Islamabad, Winnipeg, Kyoto, Hyderabad, stockholm \\
 Nigeria &  Winnipeg, Nairobi, Maharashtra, Lagos, Johannesburg, Stuttgart, Leicester, Abu, Chicago, Tulsa \\
 Iraq &  Chicago, Hyderabad, Wilmington, Lima, Baghdad, stockholm, Kashmir, Tulsa, Belfast, singapore \\
 Laos &  Bangkok, Leicester, Chicago, Kashmir, Tulsa, stockholm, Winnipeg, Lima, Rajasthan, Johannesburg \\
 USSR &  \textcolor{teal}{\textbf{Moscow}}, NYC, Midlands, stockholm, Chicago, Cardiff, Maharashtra, Pyongyang, Boulder, Columbus \\
 Kosovo &  Kashmir, Seattle, Leicester, stockholm, Tulsa, Belfast, Mosul, vegas, Rochester, Buenos \\
 China &  \textcolor{teal}{\textbf{Beijing}}, \textcolor{teal}{\textbf{Shanghai}}, Hyderabad, Brisbane, Columbus, stockholm, Maharashtra, Amsterdam, Leicester, Hamburg \\
 Guatemala &  Greenville, Tulsa, Leicester, Buenos, Johannesburg, Kashmir, Wilmington, Lima, Chicago, Rochester \\
 Tunisia &  Johannesburg, stockholm, Hamburg, Columbus, Leicester, Tulsa, Stuttgart, Winnipeg, Cardiff, Maharashtra \\
 Denmark &  \textcolor{teal}{\textbf{Copenhagen}}, Cardiff, Leicester, Brisbane, Hyderabad, Atlanta, Saras, Chicago, Hamburg, Salem \\
 Nicaragua &  Nairobi, Bangkok, Rochester, Leicester, Amsterdam, Kerala, Maharashtra, Belfast, Winnipeg, Chicago \\
 Türkiye &  Maharashtra, München, Seattle, \textcolor{teal}{\textbf{İstanbul}}, stockholm, Jakarta, Istanbul, Toronto, Milwaukee, Kyoto \\
 Bosnia &  Hyderabad, Islamabad, Belfast, Johannesburg, Jakarta, Cardiff, Rochester, Kashmir, Leicester, Lima \\
 Netherlands &  \textcolor{teal}{\textbf{Amsterdam}}, Cardiff, Midlands, Columbus, Karachi, stockholm, Nottingham, Maharashtra, Saras, Wilmington \\
 Malaysia &  Leicester, \textcolor{teal}{\textbf{Kuala}}, Cardiff, Hamburg, Maharashtra, Baltimore, Chicago, Columbus, Johannesburg, Hyderabad \\
 Venezuela &  Wilmington, vegas, Cardiff, Maharashtra, Rochester, Brisbane, stockholm, Buenos, Lima, Tulsa \\
 Sri &  Leicester, Atlanta, Kashmir, Rajasthan, Nairobi, Cardiff, stockholm, Lima, Maharashtra, Islamabad \\
 Ireland &  \textcolor{teal}{\textbf{Dublin}}, Cardiff, Belfast, Leicester, Tehran, Johannesburg, Stuttgart, Aleppo, Bethlehem, Hyderabad \\
 Liberia &  Leicester, Winnipeg, Nairobi, Johannesburg, Chicago, Kerala, Rochester, Maharashtra, Atlanta, Greenville \\
 Afghanistan &  \textcolor{teal}{\textbf{Kabul}}, Cardiff, Islamabad, stockholm, Tulsa, Chicago, Maharashtra, Kashmir, Rajasthan, Leicester \\
 America &  Columbus, \textcolor{teal}{\textbf{Chicago}}, Belfast, Sofia, Hyderabad, \textcolor{teal}{\textbf{Seattle}}, Cardiff, Johannesburg, Maharashtra, Moscow \\
 Austria &  Cardiff, \textcolor{teal}{\textbf{Vienna}}, Hamburg, Hyderabad, Leicester, Bethlehem, Stuttgart, stockholm, Columbus, Rajasthan \\
 Scotland &  Cardiff, Glasgow, \textcolor{teal}{\textbf{Edinburgh}}, Stuttgart, stockholm, Belfast, Leicester, Columbus, Maharashtra, Lima \\
 Libya &  Chicago, stockholm, Columbus, Leicester, Aleppo, Cardiff, Mosul, Lima, Wilmington, Johannesburg \\
 Uruguay &  Buenos, Seattle, Hyderabad, Maharashtra, Hamburg, Johannesburg, Wilmington, Leicester, Columbus, Cardiff \\
 Bangladesh &  Winnipeg, Cardiff, Leicester, Maharashtra, Tulsa, Atlanta, Chicago, Bangalore, Islamabad, Kashmir \\
 Bahrain &  Leicester, Chicago, Brisbane, Kashmir, Lima, Riyadh, Dubai, Wilmington, Atlanta, Saras \\
 Pakistan &  \textcolor{teal}{\textbf{Islamabad}}, Cardiff, Jakarta, Karachi, Tulsa, Leicester, Winnipeg, Atlanta, Maharashtra, Wilmington \\
 Fiji &  Lima, Leicester, Fargo, Kashmir, Brisbane, Winnipeg, Johannesburg, Cardiff, Tulsa, Edinburgh \\
 Cambodia &  Bangkok, Tulsa, Leicester, Cardiff, stockholm, Kashmir, Johannesburg, Wilmington, Kabul, Lima \\
 Singapore &  \textcolor{teal}{\textbf{singapore}}, Chicago, Leicester, Brisbane, Hamburg, Columbus, Atlanta, Kashmir, Johannesburg, Cardiff \\
 Macedonia &  Leicester, Stuttgart, Winnipeg, Rochester, Kashmir, Johannesburg, Jakarta, Maharashtra, Budapest, Lima \\
 Mongolia &  Winnipeg, Chattanooga, Leicester, Lima, Cardiff, Kyoto, Maharashtra, Johannesburg, Rajasthan, Hamburg \\
 Peru &  \textcolor{teal}{\textbf{Lima}}, Perth, Maharashtra, Winnipeg, Leicester, Chattanooga, Seattle, Hyderabad, Nairobi, Chicago \\
 Myanmar &  Bangkok, Cardiff, Tulsa, Leicester, Winnipeg, Kashmir, Maharashtra, Kyoto, Lima, Chicago \\
 Trinidad &  Leicester, Cardiff, Maharashtra, Brisbane, Rochester, Tulsa, Winnipeg, Abu, vegas, Johannesburg \\
 Colombia &  Maharashtra, Columbus, Lima, Seattle, Rochester, Wilmington, Johannesburg, Stuttgart, Amsterdam, Hyderabad \\
 Maurit &  Winnipeg, Leicester, Johannesburg, Edinburgh, Cardiff, Chicago, Stuttgart, stockholm, Moscow, Wilmington \\
 Iran &  \textcolor{teal}{\textbf{Tehran}}, Cardiff, Lima, Kashmir, Hyderabad, Leicester, Aleppo, Chicago, Stuttgart, Hamburg \\
 India &  Indianapolis, Cardiff, Maharashtra, Chicago, Hyderabad, Leicester, Lima, Columbus, Winnipeg, stockholm \\
 Spain &  \textcolor{teal}{\textbf{Madrid}}, Hyderabad, stockholm, Spokane, Cardiff, Amsterdam, Rome, Barcelona, Dallas, Johannesburg \\
 Honduras &  Wilmington, Winnipeg, Buenos, Hamburg, Nairobi, stockholm, Johannesburg, Amsterdam, Columbus, Lima \\
 USA &  \textcolor{teal}{\textbf{NYC}}, Moscow, Columbus, Midlands, \textcolor{teal}{\textbf{Chicago}}, Sofia, Karnataka, Karachi, Cardiff, Sevilla \\
\bottomrule
\end{tabular}
}
\caption{The most influencing cities of counties in the \textit{City}$\rightarrow$\textit{Country} correlation.} 
\vspace{-2mm}
\label{tab:country_city_case}
\end{table}

\begin{table}
\centering
\small
\scalebox{0.8}{
\begin{tabular}{cc}
\toprule
Father & Influencing Mothers \\
\midrule
 Omar &  Olivia, Nora, Sara, Sofia, Naomi, Diana, Uma, Rosa, Eden, Jade \\
 Victor &  Victoria, Sofia, Maria, Savannah, Sophie, Uma, Sonia, Angela, Grace, Ivy \\
 Andre &  Angela, Sofia, Sophie, Savannah, Maria, Rebecca, Ivy, Clara, Chloe, Nina \\
 Julio &  Sofia, Chloe, Maria, Carmen, Rebecca, Ivy, Rosa, Olivia, Sonia, Savannah \\
 Enrique &  Carmen, Chloe, Rosa, Clara, Sofia, Emma, Maria, Rebecca, Fiona, Olivia \\
 Amir &  Sara, Sofia, Amelia, Eden, Mei, Nora, Uma, Bella, Victoria, Diana \\
 Xavier &  Sophie, Maria, Sonia, Olivia, Emma, Leah, Clara, Uma, Jasmine, Carmen \\
 Javier &  Carmen, Chloe, Sofia, Ivy, Maria, Jasmine, Olivia, Rosa, Fiona, Jennifer \\
 Vlad &  Elena, Sofia, Chloe, Mia, Nina, Angela, Diana, Naomi, Savannah, Clara \\
 Roberto &  Chloe, Sofia, Rosa, Carmen, Lucia, Olivia, Clara, Mei, Maria, Elena \\
 Lars &  Sophie, Clara, Maria, Nina, Ella, Sara, Harper, Savannah, Rebecca, Fiona \\
 Min &  Sonia, Mei, Angela, Eden, Clara, Chloe, Grace, Maria, Harper, Savannah \\
 James &  Grace, Fiona, Ella, Savannah, Emma, Angela, Chloe, Harper, Leah, Maria \\
 Giovanni &  Lucia, Fiona, Sofia, Savannah, Rosa, Diana, Bella, Chloe, Carmen, Mei \\
 Ivan &  Ivy, Elena, Sofia, Nina, Maria, Ada, Emma, Sophie, Savannah, Sakura \\
 Diego &  Chloe, Sofia, Maria, Rosa, Angela, Carmen, Savannah, Diana, Clara, Mei \\
 Fernando &  Maria, Rosa, Fiona, Savannah, Carmen, Angela, Sofia, Luna, Clara, Ada \\
 Ethan &  Elena, Leah, Jennifer, Emma, Jasmine, Chloe, Clara, Mei, Ada, Serena \\
 Chen &  Mei, Chloe, Grace, Nina, Eden, Harper, Sofia, Rebecca, Sakura, Sonia \\
 Gabriel &  Maria, Sophie, Eden, Leah, Sara, Grace, Chloe, Rebecca, Elena, Luna \\
 Boris &  Bella, Elena, Angela, Fiona, Nina, Ada, Sofia, Sophie, Nora, Leah \\
 Jean &  Sophie, Angela, Chloe, Maria, Naomi, Carmen, Savannah, Nina, Rebecca, Lucia \\
 Dmitry &  Sofia, Elena, Chloe, Diana, Nina, Savannah, Mia, Clara, Sakura, Ivy \\
 Ahmed &  Sara, Sofia, Sophie, Nora, Uma, Victoria, Eden, Sonia, Jennifer, Mei \\
 Wei &  Mei, Chloe, Grace, Rebecca, Mia, Sofia, Ada, Nina, Angela, Harper \\
 Ibrahim &  Sofia, Sara, Eden, Uma, Victoria, Nora, Bella, Ada, Sophie, Elena \\
 Liam &  Fiona, Emma, Mia, Chloe, Nora, Leah, Grace, Jasmine, Jade, Angela \\
 Mustafa &  Sara, Sofia, Nora, Victoria, Ada, Uma, Eden, Jade, Rosa, Elena \\
 Jorge &  Maria, Carmen, Rosa, Chloe, Sofia, Diana, Elena, Fiona, Angela, Nora \\
 Leonardo &  Clara, Sofia, Jennifer, Olivia, Chloe, Jasmine, Fiona, Rosa, Lucia, Diana \\
 Luca &  Fiona, Lucia, Sofia, Angela, Maria, Savannah, Emma, Clara, Sakura, Leah \\
 Carlos &  Carmen, Maria, Rosa, Olivia, Chloe, Sofia, Clara, Sakura, Savannah, Fiona \\
 Pedro &  Maria, Rosa, Carmen, Chloe, Olivia, Clara, Sakura, Sofia, Ivy, Ada \\
 Michel &  Sophie, Lucia, Nina, Maria, Leah, Eden, Elena, Sara, Sonia, Carmen \\
 Kai &  Mei, Maria, Nina, Angela, Chloe, Eden, Jade, Uma, Sakura, Ada \\
 Benjamin &  Leah, Eden, Bella, Rebecca, Sophie, Grace, Nina, Harper, Lucia, Victoria \\
 Noah &  Rebecca, Chloe, Nina, Nora, Eden, Naomi, Sara, Grace, Leah, Ada \\
 Ali &  Sara, Nora, Eden, Victoria, Uma, Sofia, Mei, Jade, Bella, Sonia \\
 Levi &  Chloe, Leah, Eden, Sara, Nina, Elena, Harper, Bella, Rosa, Rebecca \\
 Antonio &  Rosa, Maria, Angela, Lucia, Sofia, Chloe, Savannah, Olivia, Carmen, Fiona \\
 Rafael &  Sofia, Rosa, Carmen, Maria, Clara, Leah, Ivy, Chloe, Naomi, Lucia \\
 Marco &  Maria, Sofia, Jasmine, Lucia, Clara, Angela, Chloe, Mei, Rebecca, Carmen \\
 Stefan &  Elena, Fiona, Angela, Savannah, Clara, Sophie, Mei, Maria, Eden, Rebecca \\
 Chung &  Mei, Chloe, Grace, Maria, Angela, Sonia, Harper, Clara, Savannah, Mia \\
 Abdul &  Uma, Sara, Sofia, Nora, Jennifer, Ada, Rosa, Victoria, Eden, Bella \\
 Muhammad &  Sofia, Sara, Victoria, Mei, Emily, Jennifer, Nora, Uma, Eden, Naomi \\
 Hugo &  Maria, Sophie, Chloe, Clara, Fiona, Emma, Savannah, Angela, Carmen, Ivy \\
 Axel &  Sophie, Angela, Rebecca, Nina, Ada, Emma, Fiona, Ivy, Eden, Savannah \\
 Lucas &  Lucia, Maria, Clara, Fiona, Uma, Chloe, Harper, Savannah, Sophie, Jasmine \\
 Mason &  Harper, Leah, Jasmine, Chloe, Angela, Nina, Ada, Sofia, Ella, Emma \\
 Hassan &  Sara, Eden, Nora, Victoria, Bella, Sofia, Naomi, Savannah, Mei, Diana \\
 Pablo &  Maria, Chloe, Sofia, Rosa, Savannah, Rebecca, Carmen, Elena, Fiona, Luna \\
 Raphael &  Rebecca, Sophie, Elena, Leah, Rosa, Grace, Eden, Fiona, Clara, Sonia \\
 Elijah &  Elena, Eden, Rebecca, Chloe, Savannah, Ella, Leah, Emily, Grace, Uma \\
 Louis &  Sophie, Nina, Savannah, Grace, Rosa, Maria, Rebecca, Fiona, Leah, Sonia \\
 Ricardo &  Chloe, Carmen, Sofia, Rosa, Jennifer, Clara, Rebecca, Sakura, Mei, Olivia \\
 Samuel &  Sonia, Savannah, Leah, Eden, Rebecca, Sophie, Grace, Ada, Emma, Clara \\
 William &  Grace, Emma, Emily, Leah, Ada, Harper, Angela, Victoria, Fiona, Diana \\
 Salman &  Sonia, Sofia, Nora, Uma, Sara, Bella, Eden, Jennifer, Victoria, Leah \\
 Oliver &  Olivia, Sophie, Harper, Elena, Nina, Maria, Grace, Diana, Emma, Nora \\
 Angelo &  Angela, Sofia, Fiona, Clara, Chloe, Rosa, Carmen, Savannah, Lucia, Nina \\
 Hans &  Sophie, Rebecca, Angela, Savannah, Eden, Ella, Clara, Maria, Uma, Mei \\
 Jamal &  Sofia, Jasmine, Uma, Sara, Mei, Eden, Naomi, Victoria, Bella, Diana \\
 Santiago &  Sofia, Maria, Rosa, Carmen, Chloe, Savannah, Mei, Olivia, Ivy, Luna \\
\bottomrule
\end{tabular}
}
\caption{The most influencing fathers of mothers in the \textit{Mother}$\rightarrow$\textit{Father} correlation.} 
\vspace{-2mm}
\label{tab:parent_case}
\end{table}

\begin{table}
\centering
\small
\scalebox{0.8}{
\begin{tabular}{lcc}
\toprule
& Attribute & Influencing Objects \\
\midrule
\multirow{20}*{\rotatebox{90}{Genre}}&  toys &  toy, puzzle, drum, shoes, sweater, electric, fridge, gloves, chair, jeans \\
&  transport &  headphones, pen, plate, drum, electric, car, couch, smartphone, rug, suitcase \\
&  kitchen &  drum, jeans, pen, plate, toy, backpack, rug, fridge, chair, grill \\
&  furniture &  drum, chair, fridge, electric, rug, camera, puzzle, shoes, sweater, plate \\
&  decor &  drum, rug, vase, pen, sweater, jeans, smartphone, backpack, washing, speaker \\
&  accessories &  drum, shoes, plate, laptop, electric, oven, gloves, curtains, jeans, chair \\
&  sports &  basketball, pen, drum, jeans, plate, skateboard, tennis, rug, charger, puzzle \\
&  travel &  pen, drum, water, yoga, suitcase, sunglasses, watch, plate, jeans, fridge \\
&  art &  drum, puzzle, pen, scarf, water, camera, couch, toy, chair, jeans \\
&  fitness &  yoga, puzzle, drum, pen, couch, electric, sweater, scarf, rug, camera \\
&  outdoors &  drum, plate, pen, fishing, electric, water, couch, camera, toy, puzzle \\
&  bags &  drum, fridge, sweater, gloves, jeans, backpack, pen, rug, electric, umbrella \\
&  electronics &  electric, drum, headphones, plate, toy, pen, laptop, jeans, sweater, couch \\
&  clothing &  drum, sweater, electric, shoes, skateboard, pen, jeans, camera, rug, fridge \\
&  food &  fridge, drum, pen, water, scarf, couch, plate, smartphone, sweater, speaker \\
&  photography &  camera, water, drum, puzzle, scarf, skateboard, yoga, headphones, rug, couch \\
&  literature &  book, iron, pen, drum, yoga, couch, water, speaker, scarf, fan \\
&  appliances &  electric, sweater, jeans, plate, shoes, fridge, drum, chair, oven, laptop \\
&  home &  electric, oven, drum, smartphone, pen, backpack, rug, jeans, fridge, puzzle \\
&  music &  guitar, drum, headphones, scarf, basketball, pen, toy, puzzle, suitcase, water \\
\midrule
\multirow{4}*{\rotatebox{90}{Heat}}&  warm &  hoodie, sweater, clock, lamp, drum, earrings, yoga, apple, tennis, oven \\
&  hot &  hoodie, puzzle, tennis, drum, oven, jeans, car, lamp, earrings, fan \\
&  neutral &  jeans, speaker, blanket, sofa, car, puzzle, earrings, hoodie, tennis, rug \\
&  cold &  hoodie, car, earrings, fan, lamp, curtains, couch, clock, puzzle, sweater \\
\midrule
\multirow{3}*{\rotatebox{90}{Size}}&  large &  smartphone, jeans, drum, puzzle, hoodie, umbrella, pencil, clock, car, backpack \\
&  medium &  hoodie, tripod, car, keyboard, drum, suitcase, smartphone, basketball, curtains, bottle \\
&  small &  smartphone, hoodie, car, drum, pencil, jeans, backpack, keyboard, puzzle, toy \\
\midrule
\multirow{14}*{\rotatebox{90}{Color}}&  black &  jeans, iron, fan, umbrella, hoodie, suitcase, puzzle, bowl, printer, electric \\
&  green &  backpack, plate, puzzle, jeans, couch, umbrella, drum, soap, car, sweater \\
&  blue &  jeans, electric, puzzle, plate, backpack, fishing, bottle, chair, car, umbrella \\
&  beige &  jeans, soap, hoodie, drum, puzzle, bottle, suitcase, oven, bed, speaker \\
&  gold &  puzzle, backpack, car, earrings, iron, bottle, drum, jeans, plate, fan \\
&  natural &  jeans, bottle, puzzle, earrings, car, plate, oven, yoga, suitcase, drum \\
&  silver &  bottle, jeans, puzzle, iron, drum, mirror, soap, electric, backpack, earrings \\
&  orange &  puzzle, car, drum, backpack, jeans, umbrella, bottle, electric, oven, plate \\
&  red &  car, drum, earrings, puzzle, microwave, pen, umbrella, bowl, electric, backpack \\
&  gray &  jeans, soap, mouse, puzzle, plate, sweater, umbrella, printer, bed, backpack \\
&  brown &  soap, iron, puzzle, sweater, umbrella, backpack, speaker, drum, hoodie, couch \\
&  yellow &  plate, yoga, car, backpack, umbrella, soap, drum, puzzle, sweater, fan \\
&  purple &  puzzle, drum, electric, hoodie, backpack, jeans, microwave, mouse, bottle, bowl \\
&  white &  plate, suitcase, fan, jeans, puzzle, backpack, soap, umbrella, sweater, drum \\
\midrule
\multirow{2}*{\rotatebox{90}{Price}}&  high &  smartphone, drum, air, car, hoodie, jeans, backpack, umbrella, puzzle, electric \\
&  low &  drum, jeans, backpack, smartphone, car, hoodie, air, umbrella, puzzle, electric \\
\bottomrule
\end{tabular}
}
\caption{The most influencing objects of attributes in the simile correlation.} 
\vspace{-2mm}
\label{tab:simile_case}
\end{table}

\clearpage

\section{Low Dispersion in Label-wise Correlation}

\begin{figure}
    \centering
    \caption{The std of correlation distribution between logits.}
    \label{fig:std}
    \includegraphics[width=0.81\linewidth]{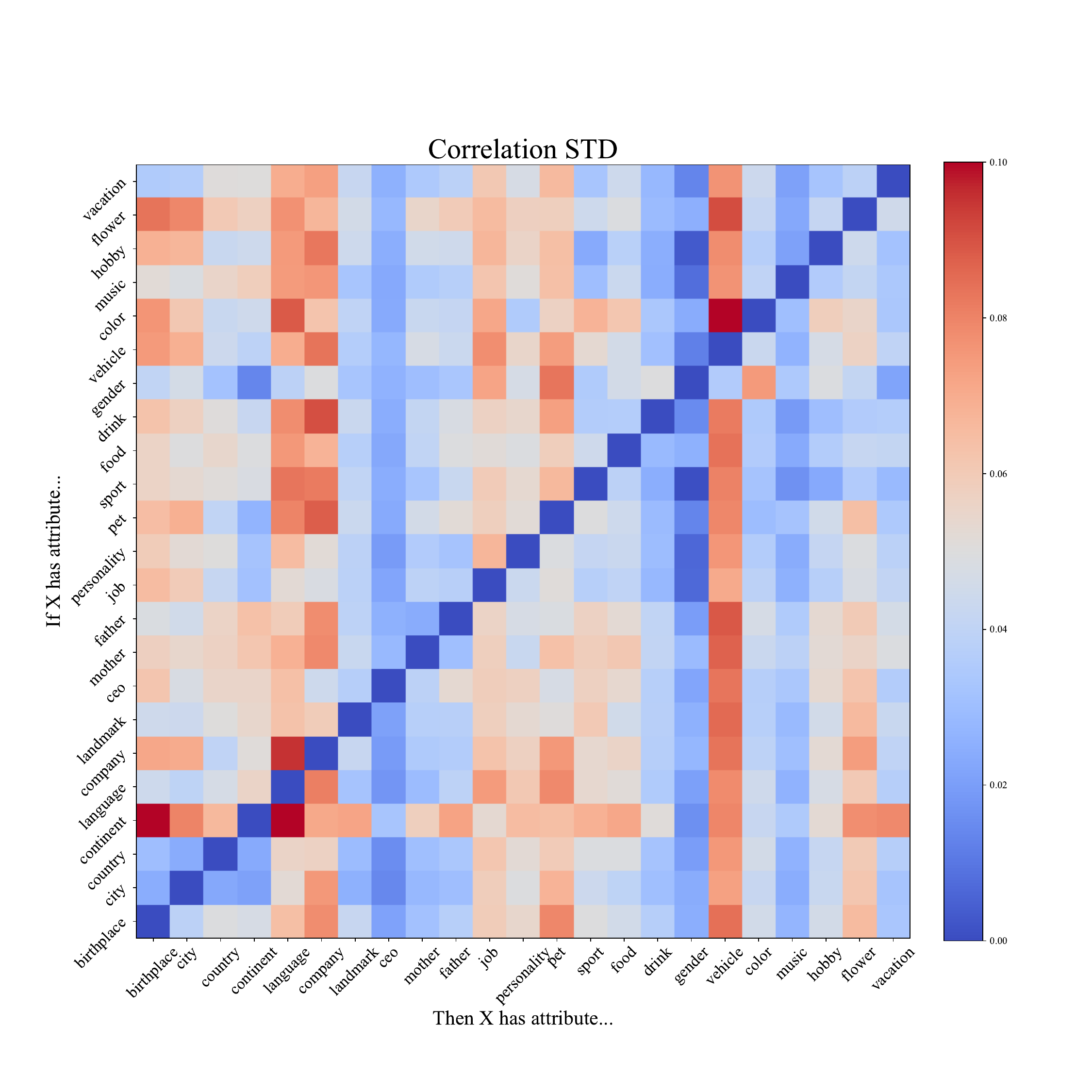}
    \vspace{-5mm}
\end{figure}

\begin{figure}
    \centering
    \caption{The std of correlation distribution between logits \textbf{before and after large-scale post-training}.}
    \label{fig:crosstuned_std}
    \includegraphics[width=0.81\linewidth]{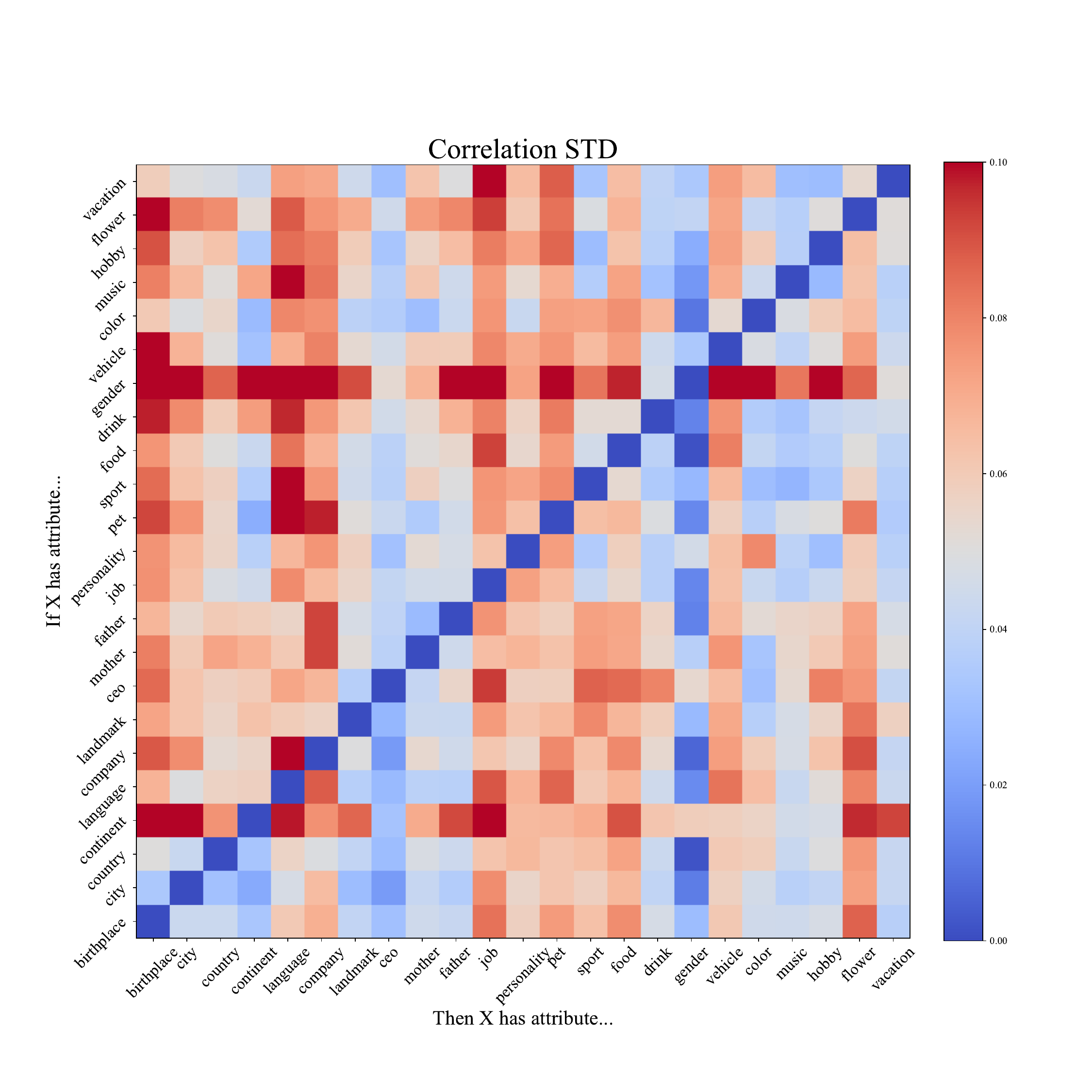}
    \vspace{-5mm}
\end{figure}

A potential concern on the correlation metric is whether the correlation reflects the majority property of different labels or some highly correlated cast bias into the evaluation. We plot the std of label-wise correlation distributions of \texttt{llama-3-8b} in Figures~\ref{fig:std} (on the same model) and~\ref{fig:crosstuned_std} (before and after post-training). The result shows the distributions to be concentrated with a std generally lower than $0.05$, which addresses the misrepresentation concern. 

\clearpage

\section{Subword Issue}
\label{apdx:subword}

Finally, we show the precision of $W$ is highly affected by the semantics of the input and output tokens. We first categorize the tokens into $3$ categories, 1) Subword, a token being part of a word, such as a prefix like \textit{Br} in \textit{Brunei}, 2) Word in a phrase, a token is a whole word but also a part of a phrase like \textit{North} in \textit{North America}, 3) Whole semantics, the rest of tokens with a full meaning in itself like \textit{USA}. 

\begin{table}
\centering
\small
\scalebox{0.95}{
\begin{tabular}{cccc}
\toprule
Completeness & Correlation & Precision (Hit@Top-$5$) & Generalization \\
\midrule
Whole Semantics & $0.85$ & $0.49$ & $55.67\%$ \\
Word in a Phrase & $0.86$ & $0.10$ & $2.00\%$ \\
Subword & $0.87$ & $0.00$ & $0.00\%$ \\
\bottomrule
\end{tabular}
}
\caption{The correlation and $W$ precision of tokens with different levels of semantic completeness.} 
\vspace{-2mm}
\label{tab:subword}
\end{table}

The results in Table~\ref{tab:subword} show the semantic completeness to be an important factor in whether knowledge can be generalized. With higher semantic completeness (Whole Semantics $>$ Word in a Phrase $>$ Subword), the $W$'s precision also rise as the token indicates a clearer entity. Consequently, it can be better updated by the generalization behavior caused by the linear correlation. The only precise mapping (and successful) generalization for ``Word in a Phrase'' is \textit{Riyadh}$\rightarrow$\textit{Saudi Arabia}, where the first token \textit{Saudi} has a strong indication of the country. 


\end{document}